\documentclass[runningheads]{llncs}
\usepackage{graphicx}
\usepackage{multirow}

\usepackage{tikz}
\usepackage{comment}
\usepackage{amsmath,amssymb} 
\usepackage{color}
\usepackage{bm}
\usepackage{makecell}
\usepackage{boldline}
\usepackage[accsupp]{axessibility}  

\newcolumntype{?}[1]{!{\vrule width #1}}

\begin{document}
\pagestyle{headings}
\mainmatter
\def\ECCVSubNumber{4323}  

\title{Contrastive Learning for Diverse Disentangled Foreground Generation
} 

\titlerunning{ContrasFill}

\author{Yuheng Li\inst{1,2} \and
        Yijun Li\inst{2} \and
Jingwan Lu\inst{2} \and
Eli Shechtman\inst{2} \and
Yong Jae Lee\inst{1} \and
Krishna Kumar Singh\inst{2}
}
\authorrunning{Li et al.}

\institute{$^{1}$University of Wisconsin-Madison $^{2}$Adobe Research}

\maketitle

\begin{abstract}
We introduce a new method for diverse foreground generation with explicit control over various factors. Existing image inpainting based foreground generation methods often struggle to generate diverse results and rarely allow users to explicitly control specific factors of variation (e.g., varying the facial identity or expression for face inpainting results). We leverage contrastive learning with latent codes to generate diverse foreground results for the same masked input. Specifically, we define two sets of latent codes, where one controls a pre-defined factor (``known''), and the other controls the remaining factors (``unknown''). The sampled latent codes from the two sets jointly bi-modulate the convolution kernels to guide the generator to synthesize diverse results. Experiments demonstrate the superiority of our method over state-of-the-arts in result diversity and generation controllability. 

\end{abstract}

\section{Introduction}
\label{sec:intro}

\begin{figure}[t!]
    \centering
    \includegraphics[width=0.95\textwidth]{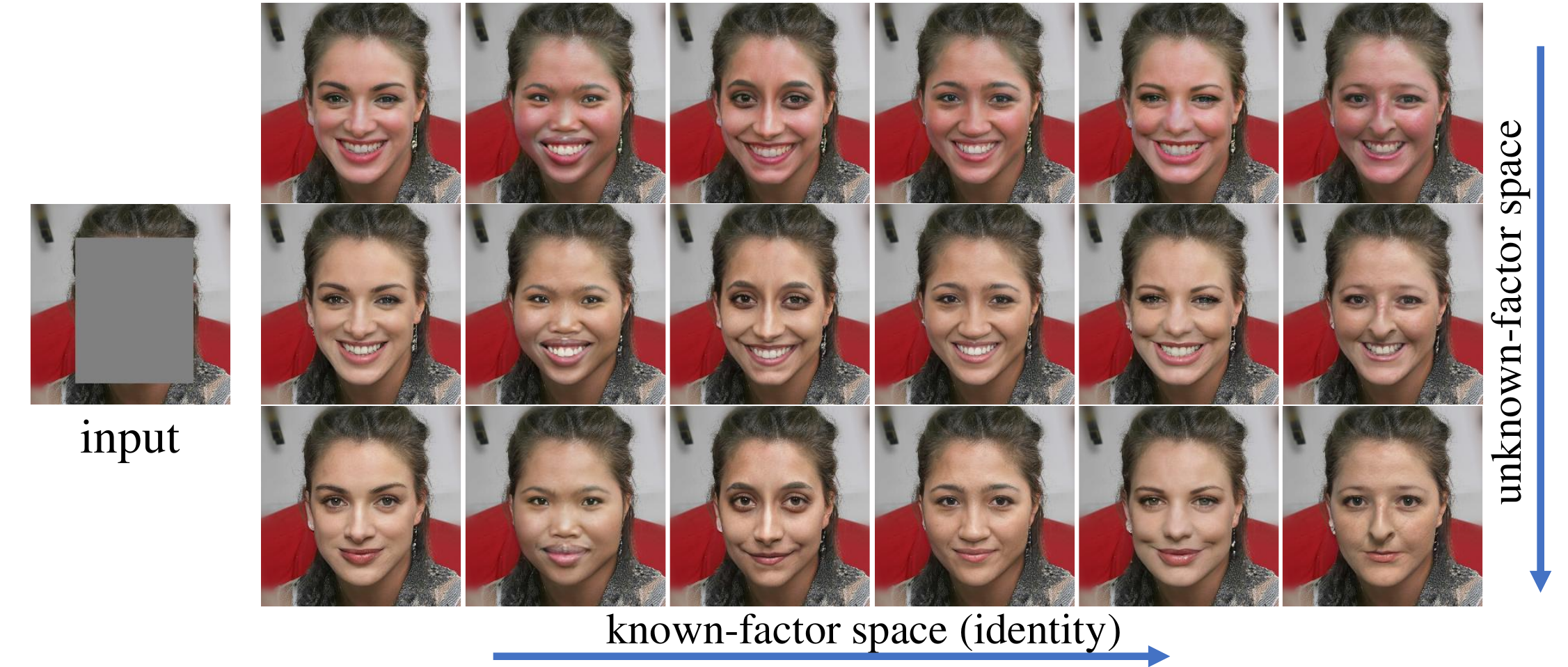}
    \caption{\textbf{Foreground generation on the same mask.} We use contrastive learning to increase generation diversity. We also explicitly disentangle out an expected predefined factor (human identity here) to increase diversity and controllability.}
    \label{fig:teaser}
\end{figure}

Foreground object generation is the task of filling in the missing foreground region in a given context, such as generating human faces as shown in Figure~\ref{fig:teaser}. This task is useful in practice, e.g., for privacy-related applications (anonymizing a person's face by generating a new identity) or replacing/adding objects in an image (replacing a car in a photo if one does not like the original one). It is a special case of image inpainting in which the entire foreground object is masked. In inpainting, when the missing region (hole) is small, there may only be one or few ``correct'' completions (e.g., if only one eye is masked, then it mostly can be inferred from the other eye), but as the hole gets bigger there should be more diversity in the generated completion, especially when an entire object is masked. As there can be many different plausible solutions for filling in the missing region, this task naturally demands learning a ``one-to-many'' mapping between the input and outputs (e.g., Figure~\ref{fig:teaser}). That is, a good method should 1) synthesize foreground objects that are both \emph{realistic} and \emph{semantically coherent} with the surrounding unmasked context; 2) have the capability to generate \emph{diverse} results for the same missing region and context; and 3) provide \emph{control} over different properties of the synthesized results. While tremendous progress has been made to obtain better realism and coherence~\cite{Zhao2021LargeSI,Guo_2021_ICCV,Suin_2021_ICCV,Li2021CollagingCG}, progress in diversity is still unsatisfactory and increasing controllability for the results is also relatively under explored.         

Like the inpainting task~\cite{Marcelo-2000,Hays2007SceneCU,Barnes2009PatchMatchAR,Pathak2016ContextEF,Liu2018ImageIF,Zheng2019PluralisticIC}, foreground generation needs to consider coherence between the given context and the generated object. Existing inpaiting work can generate good quality object/foreground, but it usually lacks diversity and controllability. Although there are many inpainting methods trying to generate diverse results~\cite{Zheng2019PluralisticIC,Zhao2020UCTGANDI,Zhao2021LargeSI,Li2021CollagingCG}, the results are still less satisfactory. These methods typically have an encoder-decoder architecture. To achieve diversity, different latent codes can be sampled and injected into these models. However, although the output is a function of both the masked image as well as the latent code, the spatial features from the encoder usually dominate the final results and prevent the latent codes from inducing large changes. For example, in \cite{Zhao2021LargeSI}, an encoder is used to extract 2D spatial features from the masked image, and skip connections are added to all levels of the encoder and decoder. The information from the latent code can be easily submerged by the large number of features from the encoder.

In this paper, we propose a novel approach for diverse and controllable foreground generation. As shown in Figure~\ref{fig:teaser}, our method can generate diverse results for the same input. To synthesize diverse content, we condition the generation on both the masked image and the sampled latent codes, and apply contrastive learning~\cite{Chen2020ASF} so that the latent codes that are close/far in code space result in corresponding synthesized images that are close/far in image space.

Besides diversity, controllability is another desired property in foreground generation. Thus, we also try to explicitly disentangle a predefined factor by using a pretrained classifier on this factor. For example, as shown in Figure~\ref{fig:teaser}, one can disentangle human identity (rows) from other attributes (columns) for face images. We explicitly use two sets of latent codes, where one represents the predefined factor (``known''), and the other controls all the other factors (``unknown''). This allows us to change the unknown factors while keeping the known factor fixed (e.g,, in Figure~\ref{fig:teaser}, changing the facial attributes which are unknown during training  while keeping the identity of the face intact). To inject these two codes, we propose a bi-modulated convolution module where the convolution kernels are modulated by the two latent codes from different spaces. We design each training batch to contain a mix of instances that share the same known latent code while differing in the unknown, and instances that share the same unknown latent code while differing in the known. We use a contrastive loss to ensure that known and unknown codes control their respective factors.

\noindent\textbf{Contributions.} (1) We propose a novel contrastive learning based approach for diverse foreground generation; (2) An explicit disentangled latent space for controllability via a novel bi-modulated convolution module; (3) More diverse results compared to existing state-of-the-art methods on three different datasets.

\section{Related Work}

\noindent\textbf{Image inpainting} This problem has been studied for decades due to its importance. Traditional methods~\cite{935036,Marcelo-2000,Telea-2004,Hays2007SceneCU,Barnes2009PatchMatchAR} typically rely on low-level assumptions and image statistics, leading to over smoothing and results with limited visual semantics. Recently, deep learning methods~\cite{Ding2019ImageIU,Iizuka2017GloballyAL,Ren2019StructureFlowII,Sagong2019PEPSIF,Wang2018ImageIV,Xie2019ImageIW,Xiong2019ForegroundAwareII,Yan2018ShiftNetII,Yang2017HighResolutionII,Zeng2019LearningPE,Liu2018ImageIF,Yu2019FreeFormII,Zhao2020UCTGANDI} dramatically boosted the quality, in terms of both visual quality and semantic coherence. \cite{Pathak2016ContextEF} first uses an encoder-decoder architecture in inpainting with reconstruction loss and adversarial loss~\cite{goodfellow-nips2014}. \cite{Liu2018ImageIF} and~\cite{Yu2019FreeFormII} proposes the use of partial and gated convolutions on irregular masks. However, these methods only generate deterministic results. Thus~\cite{Zheng2019PluralisticIC} proposes a VAE-based~\cite{Kingma2014AutoEncodingVB} method allowing pluralistic image completion. Recently proposed~\cite{Zhao2021LargeSI,Li2021CollagingCG} use StyleGAN~\cite{Karras-CVPR2018,karras-cvpr2019} architecture for inpainting. \cite{Zhao2021LargeSI} combines encoded features from a masked image with a random latent code to co-modulate StyleGAN convolution kernels. \cite{Li2021CollagingCG} has a similar setting as ours as instead of traditional inpainting, they use a foreground model to synthesize high quality foreground objects conditioned on the background context. In both work, diverse images can be generated by sampling different latent codes injected into StyleGAN. But their diversity in the latent code space is restricted due to extra spatial features from the encoder which usually determine most of the aspects of the generation. \cite{Wan2021HighFidelityPI,Yu2021DiverseII} also try to use transformer to realize diversity in image inpainting. They both use bidirectional attention to predict missing tokens. However, their image quality suffers compared to styleGAN2-based architectures. Also, none of the existing pluralistic inpainting work enables user controllability in the results via latent code disentanglement.          

\noindent\textbf{Contrastive learning} Contrastive learning~\cite{Zhuang2019LocalAF,Tian2020ContrastiveMC,He2020MomentumCF,Misra2020SelfSupervisedLO,Chen2020ASF} has shown great potential in representation learning. Among them, \cite{Chen2020ASF} proposes a simple framework for contrastive learning without requiring specialized architectures or memory bank. Recently \cite{Park2020ContrastiveLF} proposes to use contrastive learning in image translation task. Also, there are a few work~\cite{Zhou2021ImageIW,Ma2021FreeFormII} studying contrastive learning in the image inpainting task. Like most inpainting methods, \cite{Zhou2021ImageIW,Ma2021FreeFormII} use encoder-decoder architecture. \cite{Zhou2021ImageIW} encode more discriminative features using contrastive loss in different semantic sub-regions. \cite{Ma2021FreeFormII} applies the contrastive loss to the output features of encoder by setting two identical images with different masks as positive pairs while different images as negative pairs. However, they are both deterministic inpainting methods which means they only produce a single result per input. Different from prior work, instead of learning a better intermediate feature using contrastive loss, we use contrastive learning to achieve disentanglement and diversity in the latent space, enabling us to produce diverse inpainting results for foreground generation in a controllable way.  

\noindent\textbf{Disentanglement learning} For a generative model, it is desirable to disentangle the factors of variation. One way is to explicitly learn a disentangled latent space: having separate codes for different factors. A large number of work try to disentangle object shape/structure from appearance~\cite{Singh2019FineGANUH,Li2020MixNMatchMD,Shu2018DeformingAU,Denton2017UnsupervisedLO,Xing2020DeformableGN}. Searching for semantic directions in a pre-trained GAN latent space is another way to achieve disentanglement. This method is getting popular recently and both unsupervised methods~\cite{Voynov2020UnsupervisedDO,harkonen2020ganspace,Shen2020ClosedFormFO} and supervised methods~\cite{gansteerability,Shen2020InterpretingTL,Yang2021SemanticHE} are heavily explored. 
Despite the progress made in the field, few work explore it for the foreground generation task. We use contrastive learning to explicitly learn a disentangled latent space for controllable foreground generation.

\section{Approach}

Our goal is to propose a model (ContrasFill) which is able to generate diverse foreground objects for the same masked region while providing control over different factors of generated results. We encode spatial features corresponding to masked image and modulate them with randomly sampled latent codes to generate diverse results. However, without applying explicit training loss on diversity, different latent codes might introduce only minor changes as in~\cite{Zhao2021LargeSI,Li2021CollagingCG}. Thus we use contrastive learning to encourage the model to synthesize diverse results by forcing latent codes closer in the latent space to produce images closer in the image space and vice versa. To gain explicit control over certain factor, we also try to disentangle the latent space into two spaces: a known factor space which corresponds to an expected factor, and an unknown factor space which controls the rest of other factors.    
Section~\ref{sec:Contrastive_learning_for_diversity_and_disentanglement} introduces the training details for achieving diversity and disentanglement using the contrastive loss. Section~\ref{sec:Codes_injection_with_bi-modulated_convolution} talks about how do we inject two codes into our model using the proposed bi-modulation.

\begin{figure*}[t!]
    \centering
    \includegraphics[width=0.95\textwidth]{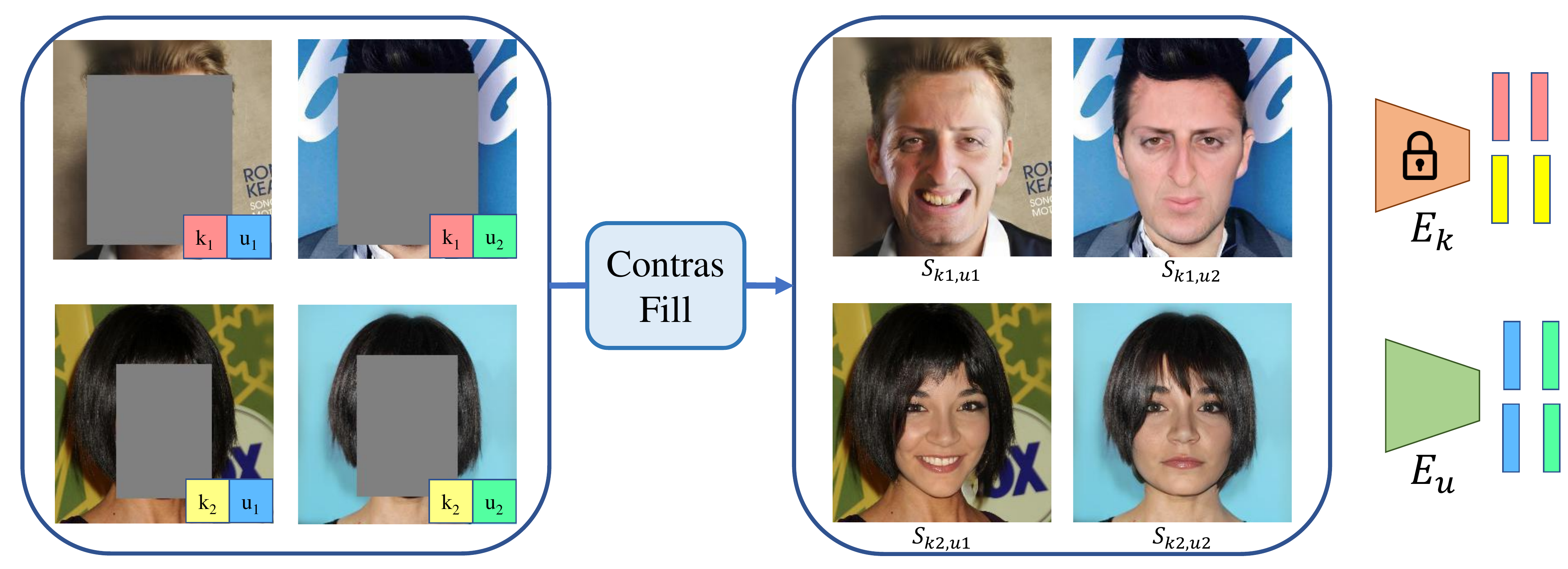}
    \caption{ContrasFill takes as input two sets of codes (squares on the left): known-factor code (e.g., identity) and unknown-factor code (non-identity factors) to synthesize images. Two encoders ($E_k$ and $E_u$) embed images into different features (bars on the right, color between code and feature refers to correspondence). A contrastive loss forces features with same/different colors closer/further in the feature spaces.}
    \label{fig:apprach}
\end{figure*}

\subsection{Contrastive learning for diversity and disentanglement}
\label{sec:Contrastive_learning_for_diversity_and_disentanglement}
Figure~\ref{fig:apprach} shows the framework of our method. Our model takes a masked image $I$ with context only and two latent codes as inputs: a known factor code $k$ drawn from a distribution $\phi_k$ and an unknown factor code $u$ from a distribution $\phi_u$ to output the synthesized image $S$. If we define our model as $G$, then we have $S_{k,u,I} = G(k, u, I)$.  Since the input context is independent of the following analysis, we will omit $I$ in $S_{k,u,I}$ for simplicity and talk about the context later. One can refer to supp. for visual examples to help understand the following analysis.  

Suppose we sample $N$ codes from each latent space, we will have $N^2$ combinations between the code $k$ and $u$ in total. To enforce a contrastive loss in the known and unknown factor latent spaces, we first define an image pair as: 
\begin{equation}
\label{eq:image_pair}
p_{(k,u), (k',u')} = (S_{k,u}, S_{k',u'}), 
\end{equation}
Note, we do not consider images sharing the same known and unknown codes as a valid pair. 
In order words, $k=k'$ and $u=u'$ cannot hold at the same time. 
We will next define the positive and negative pairs used in our contrastive learning scheme. To simplify the explanation, we first consider the case of the known space.

\noindent\textbf{Contrastive pairs in the known space.} A positive image pair contains two images sharing the same known codes but different unknown codes (i.e., $k=k', u \neq u'$ in the Eq~\ref{eq:image_pair}). We define $P_{ \textbf{k},u}$ as a set of all positive pairs \emph{associated} with the code combination $(k,u)$ in the known space (bold indicates the space). For example, in  Figure~\ref{fig:apprach} where we set the known factor as human identity, $P_{\textbf{k1},u1} = \{  p_{(k1,u1), (k1,u2)} \}$.  To ease explanation, we denote $P_{ \bm{K} }$ as \emph{all} positive pairs in the known space. Here we have $ P_{ \bm{K} } = \{  p_{(k1,u1), (k1,u2)}, p_{(k2,u1), (k2,u2)} \}$. 

For the negative pair, we define two images not sharing the same known codes (i.e., $k \neq k'$). In this case, we construct two types of negative pairs. The first case is the hard negative pair where two images sharing different known codes but the same unknown codes (i.e., $k \neq k', u = u'$, images in each column in Figure~\ref{fig:apprach}). The reason is that these images share the same features (e.g., smile) in the unknown space which forces the learned known latent code to control different aspects of the face due to the use of contrastive loss which we will introduce later. Pairs of images sharing different known and unknown codes (i.e., $k \neq k', u \neq u'$) are easy negative pairs (diagonal image pairs in Figure~\ref{fig:apprach}). Similarly, we define $N_{\textbf{k},u}$ as all negative pairs \emph{associated} with the code combination $(k,u)$ in the known factor space. For example, for image $S_{k1,u1}$ in Figure~\ref{fig:apprach}, $N_{\textbf{k1},u1} = \{ p_{(k1,u1), (k2,u1)}, p_{(k1,u1), (k2,u2)} \} $.

\noindent\textbf{Contrastive loss in the known space.} The job of the known space is to control an expected factor during the generation. To push the model to learn this correspondence, we use contrastive learning. The intuition is to push images closer/further if they are positive/negative pairs. In order to measure the distance between two images, we define the similarity score $f$ as: 

\begin{equation}
\label{eq:sim}
f_{(k,u), (k',u')} = e^{\mathrm{sim}(\bm z_{k,u}, \bm z_{k',u'})/\tau},
\end{equation}
where $\bm z_{k,u}$ is the extracted feature of the image $S_{k,u}$ from an encoder, $\mathrm{sim}(\cdot, \cdot)$ is the cosine similarity and $\tau$ denotes a temperature parameter. To force our known space to control the expected factor, we assume having access to a pretrained and fixed classifier. The encoder $E_k$ for the known space will output the penultimate feature of the classifier. For example, in faces, we use a pretrained ArcFace~\cite{Deng2019ArcFaceAA} to extract identity features.

For an image $S_{k,u}$, and its positive pair $S_{k,u'}$ in the known space, the contrastive loss becomes:

\begin{equation}
\label{eq:known_pair_contrastive}
\ell_{(k,u), (k,u')} = -\log \frac{ f_{(k,u), (k,u')} }  { f_{(k,u), (k,u')} + FN_{\textbf{k},u}}, 
\end{equation}
where $FN_{\textbf{k},u}$ is the sum of similarity scores of all negative pairs with respect to the image $S_{k,u}$. In other words, it is the summation of Eq.~\ref{eq:sim} over all elements in the $N_{\textbf{k},u}$.
Finally, the total loss for the known space becomes: 
\begin{equation}
\label{eq:known_loss}
\mathcal{L}_{known} = \frac{1}{ | P_{ \bm{K} }| } \sum  \ell_{(k,u), (k,u')}, 
\end{equation}
where the summation is over all positive pairs $ P_{ \bm{K} }$ in the known space.

\noindent\textbf{Contrastive learning in the unknown space.} The contrastive learning idea is similarly applied to the unknown latent space and we highlight the main difference below.

In the unknown space, positive pairs share the same unknown code (each column in Figure~\ref{fig:apprach}) and negative pairs have different unknown codes. Similarly, we define $P_{\bm{U}}$ as \emph{all} positive pairs in the unknown space.  For example, in Figure~\ref{fig:apprach}, $P_{\bm{U}} = \{  p_{(k1,u1), (k2,u1)}, p_{(k1,u2), (k2,u2)} \}$. For an image  $S_{k,u}$, we define all negative pairs associated with it in the unknown space as $N_{k,\textbf{u}}$ (bold indicates space). For example,  $N_{k1, \textbf{u1}} = \{ P_{(k1,u1), (k1,u2)}, P_{(k1,u1), (k2,u2)} \} $  In the unknown space, the image feature $\bm z_{k,u}$ for calculating image pair similarity in the Eq~\ref{eq:sim} is extracted from an encoder $E_u$ which is trained from scratch. This is because it is hard to define what factors can be controlled in the unknown space beforehand. Also, this avoids pre-training an additional feature extractor and simplifies our approach.

Then, for an image $S_{k,u}$, and its positive pair $S_{k',u}$, the counterpart of Eq~\ref{eq:known_pair_contrastive} in the unknown space becomes

\begin{equation}
\label{eq:unknown_pair_contrastive}
\ell_{(k,u), (k',u)} = -\log \frac{ f_{(k,u), (k',u)} }  { f_{(k,u), (k',u)} + FN_{k,\textbf{u}} }, 
\end{equation}
where $FN_{k,\textbf{u}}$ is the sum of similarity score of all negative pairs with respect to image $S_{k,u}$ in the unknown space. The total loss for the unknown space becomes: 
\begin{equation}
\label{eq:unknown_loss}
\mathcal{L}_{unknown} = \frac{1}{ |P_{\bm{U}}| } \sum  \ell_{(k,u), (k',u)}, 
\end{equation}
where the summation is over all positive pairs in the $P_{\bm{U}}$ in the unknown space.

In this way the disentanglement can be learned because we use a pretrained encoder for the known-factor which only extracts expected features, thus the model will synthesize known-factors in images according to codes sampled from known space. For the unknown space, due to the existence of hard negative pair (sharing the same known factors), different unknown codes need to generate factors that are different from known factor to minimize the contrastive loss. 

Overall we have the final loss $\mathcal{L}$ as
\begin{equation}
    \mathcal{L} = \mathcal{L}_{gan} + \lambda_1 \mathcal{L}_{known} + \lambda_2\mathcal{L}_{unknown},
\label{eqn:main_loss}
\end{equation}
where $\mathcal{L}_{gan}$ is same as the one used in the StyleGAN2~\cite{Karras2019stylegan2}. $\mathcal{L}_{known}$ and $\mathcal{L}_{unknown}$ are two contrastive losses in known and unknown latent spaces. $\lambda_1$ and  $\lambda_2$ are their weights. We sample different context (background) for different code combinations (e.g., $N^2$ in total) since we want to have the same context distribution for both the real and fake batches when training the discriminator. More details are presented in the supp.

\subsection{Codes injection with bi-modulated convolution}
\label{sec:Codes_injection_with_bi-modulated_convolution}

Our model uses an encoder-decoder architecture (details in the supp). Inspired by StyleGAN2
\ that shows the effectiveness of modulation, we also use our latent codes to modulate convolution kernels. However, since we have two latent codes, we propose the bi-modulation, where the convolution kernel is modulated by two codes. We use this novel modulation scheme for all convolutions in our model.

\begin{figure}[t!]
    \centering
    \includegraphics[width=0.45\textwidth]{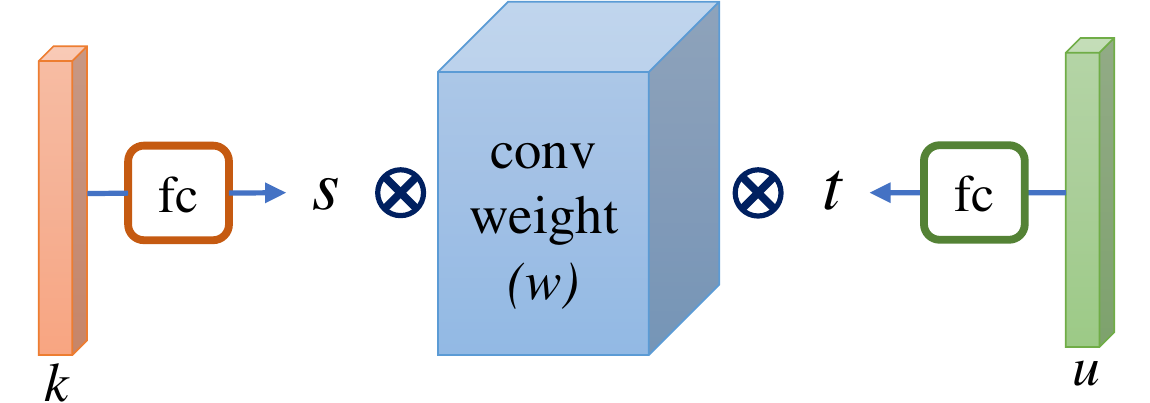}
    \caption{The proposed bi-modulation scheme, where convolution kernels are modulated by two disentangled latent codes.}
    \label{fig:archi}
\end{figure}

Figure~\ref{fig:archi} shows the bi-modulation process. The two codes $k$ and $u$ first go to two separate fully connected layer to become scaling vectors $s$ and $t$. The length of scaling vectors is the same as the number of input channels of a convolution kernel. Then the scaling vectors bi-modulate the convolution weight by: $w'_{ijk} = s_i \cdot t_i \cdot w_{ijk}$,
where $w$ and $w'$ are the original and the bi-modulated weights. $s_i$ and $t_i$ are the scaling factors corresponding to the $i$th input feature map. $j$ and $k$ enumerate the output feature maps and the spatial footprint of the convolution. 

\section{Experiments}
We perform quantitative and qualitative evaluations via comparing our proposed foreground generation model ContrasFill with prior arts.

\noindent\textbf{Datasets.} We conduct the evaluation on three different datasets: 1) \textbf{Face}. We use CelebAMask-HD~\cite{lee-cvpr2020} that includes 30,000 face images with segmentation masks. We follow the official training/testing split. To acquire more training data, we use a publicly available face parsing model~\cite{face_parsing_model} on FFHQ~\cite{karras-cvpr2019} as extra training data. We use a pretrained face recognition model~\cite{Deng2019ArcFaceAA} as our known factor feature extractor. 
2) \textbf{Bird}. We use the $bird$ category from LSUN dataset~\cite{Yu2015LSUNCO}. We choose images greater than certain resolution and run the pretrained MaskRCNN~\cite{He2020MaskR,wu2019detectron2} to remove bad images. In total, we have 34,969 images and we randomly select 10\% (3,497) as test data. We train a fine-grained classification model~\cite{du2020fine} on the CUB dataset~\cite{WahCUB_200_2011} as our known factor feature extractor.
3) \textbf{Car}. We use the $car$ category from LSUN dataset~\cite{Yu2015LSUNCO} and same preprocessing steps to clean our data. In total, we have 77,840 images and we randomly select 10\% (7,784) as test data. We train a shape classifier~\cite{du2020fine} on the Stanford car dataset~\cite{KrauseStarkDengFei-Fei_3DRR2013} as our known factor feature extractor.
To measure the extent of our ability to synthesize diverse results, we use the object bounding box as the missing region in our main study. 

We train our model at $256\times256 $ resolution on all datasets. Our unknown factor code is drawn from the normal distribution. Our known factor codes are drawn from one-hot distribution for the cars and birds; for faces, we choose to draw from a hypersphere which is the feature distribution of penultimate layer of ArcFace. We sample $N=8$ different known and unknown codes in each training minibatch. Due to memory issue, we can not fit all 64 combinations, thus we subsample one hard negative pair for each code, resulting in a batch size of 16 during training. Please refer to supp for more dataset and implementation details.

\noindent\textbf{Baselines.} We mainly compare with: 1) \textbf{CollageGAN}~\cite{Li2021CollagingCG}, which generates foreground object conditioned on the background; 2) \textbf{CoModGAN}~\cite{Zhao2021LargeSI}, a state-of-the-art image inpainting model. These two methods, built on top of StyleGAN2~\cite{Karras2019stylegan2}, are able to generate multiple results via sampling codes in the latent space; 3) \textbf{BAT-fill}~\cite{Yu2021DiverseII}, a recently proposed two-stage inpainting model using transformer. It first autoregressively predicts missing tokens in a $32 \times 32$ image with bidirectional attention, and then use a convolutional network to perform upsampling to $256 \times 256$. It samples different plausible missing tokens in the $32 \times 32$ grid to achieve diversity. This work has demonstrated the benefits of adding autoregressive predication over other transformer-based inpainting work~\cite{Wan2021HighFidelityPI}.  We train all baseline models with the same input mask setting as ours.

\noindent\textbf{Evaluation.}
We use the following metrics: 1) \textbf{FID}~\cite{Heusel2017GANsTB} measures the quality and diversity by comparing distributions between the real and the generated images. 2) \textbf{LPIPS}~\cite{zhang2018perceptual} measures distance between two images in deep feature space. For each testing image, we compare pairs of inpainting results generated from the same input mask, we use this to measure diversity. 3) Known Factor Feature Angle (\textbf{KFFA}). To better understand how we can improve result diversity using the disentangled known factor, we sample 10 inpainting results for each input image. Then we compute deep features of these 10 results from a known factor classifier. We report average angle between all normalized feature pairs. For a fair comparison, we use feature extractors different from the one used in the training. For face, we use CurricularFace~\cite{huang2020curricularface}, and for bird and car, we train a new classifier using the VGG architecture~\cite{Simonyan2015VeryDC}. Note that L1, SSIM and PSNR are also commonly used metrics for inpainting tasks. However, they all favor deterministic methods which aim to reproduce the single ground truth. As pointed out by~\cite{Wan2021HighFidelityPI}, these metrics are more suitable for small mask cases where the synthesized contents are more likely to be similar to the ground truth. With large holes covering an entire semantic region or object, synthesized diverse contents might look plausible but different from the ground truth. 

\begin{figure}[t!]
    \centering
    \includegraphics[width=0.85\textwidth]{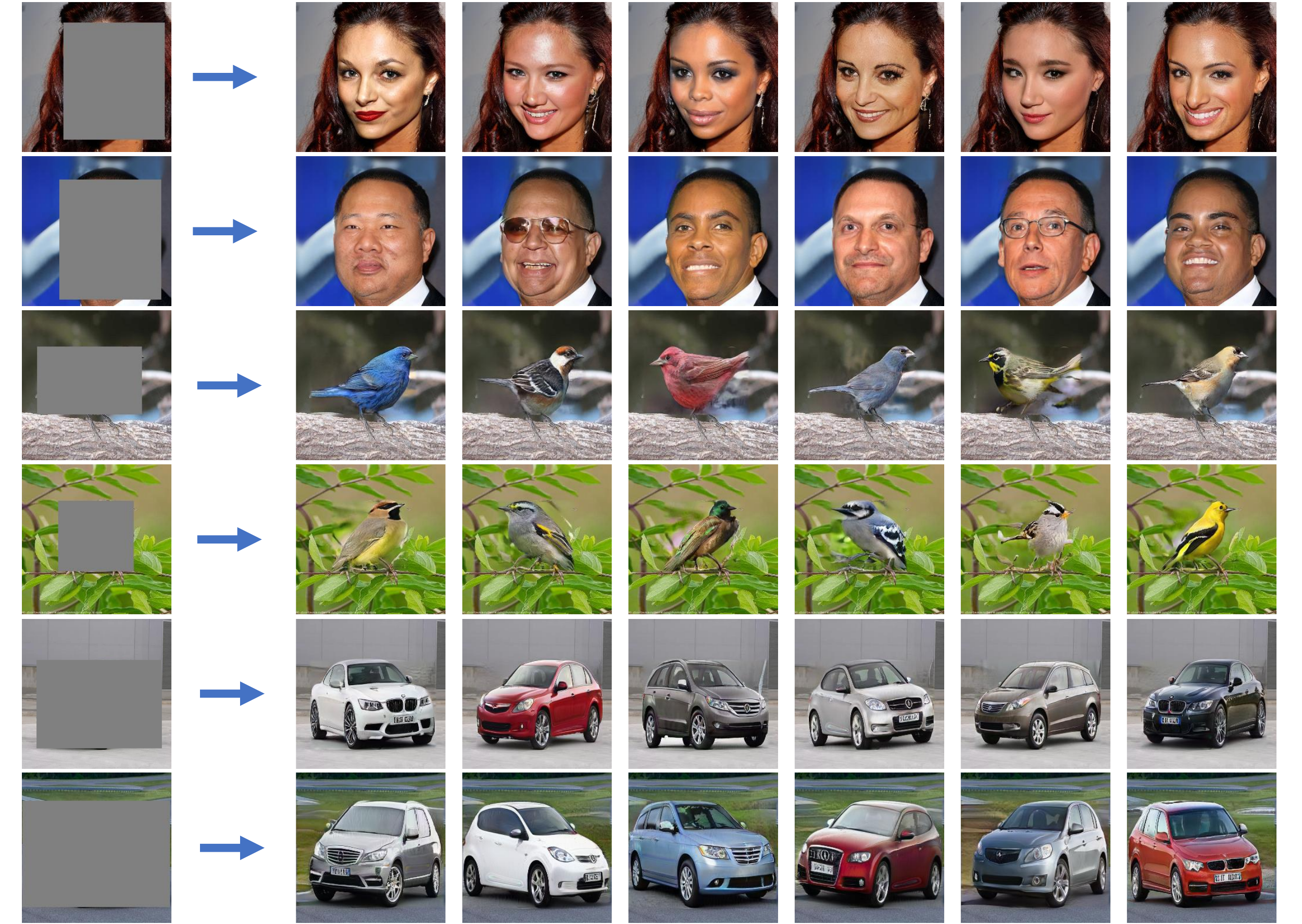}
    \caption{We can achieve diverse samples from the same masked image by sampling in both known and unknown code spaces.}
    \label{fig:diversity}
\end{figure}

\subsection{Qualitative results}

\begin{figure*}[t!]
    \centering
    \includegraphics[width=0.90\textwidth]{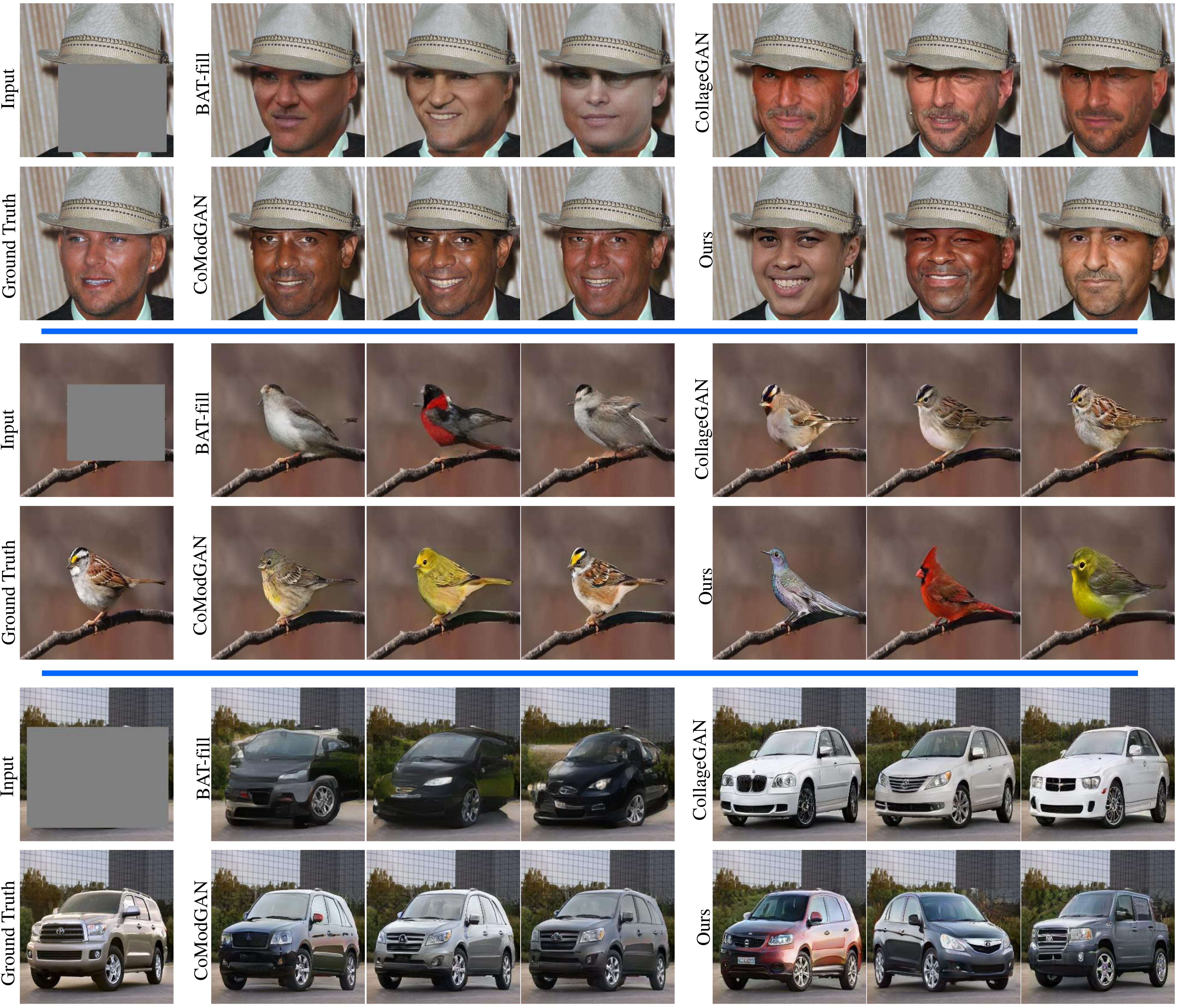}
    \caption{Compared with the baselines, our method generates more diverse results. For faces, we have more variations in identity. For birds and cars, we have different object shapes and textures.} 
    \label{fig:baseline}
\end{figure*}

Figure~\ref{fig:diversity} shows random samples from our model given the same masked input. Our method can generate diverse identities, facial attributes for faces and synthesize diverse shapes, poses and object appearances for birds and cars. 
Figure~\ref{fig:baseline} shows side-by-side comparisons between our method and other baselines. Our results on faces are more diverse compared to CoModGAN~\cite{Zhao2021LargeSI} and CollageGAN~\cite{Li2021CollagingCG}. For unaligned dataset (cars and birds), these two methods tends to generate results with the same shape. BAT-fill~\cite{Yu2021DiverseII} results have better diversity, but lower image quality. It sometimes generates artifacts on faces or distorted geometries for cars.

\begin{figure*}[t!]
    \centering
    \includegraphics[width=0.98\textwidth]{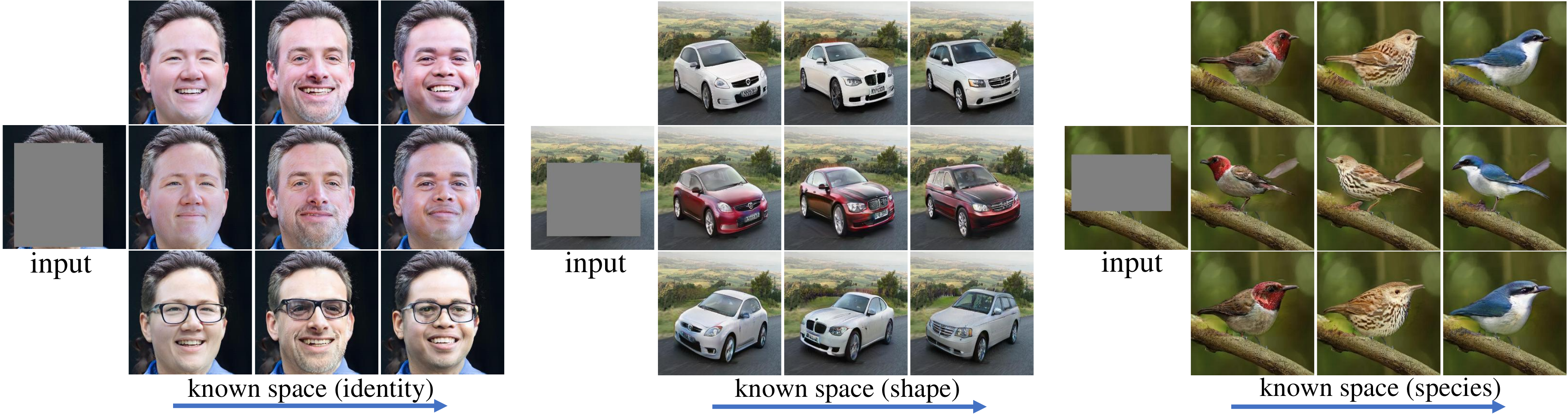}
    \caption{Known factor code is same for each column, unknown factor code is same for each row.} 
    \label{fig:disentanglement}
\end{figure*}

We also evaluate how disentangled our results are in Figure~\ref{fig:disentanglement}. Each column shares the same known factor and each row shares the same unknown factor. For faces, the known code controls identity and the unknown code controls facial attributes such as smile, glass and lighting condition. For cars, the known code controls car shape (e.g., sedan-like and wagon-like in the second and third column), and the unknown code changes color and orientation. For birds, the known factor changes species (color is associated with species) and the unknown factor changes pose and orientation. When one latent code changes, the image changes only along the direction it is supposed to, e.g, as the identities change, the same facial expression remains within each row in the face results.

\begin{table}[t!]
\footnotesize
    \begin{center}
		\begin{tabular}{ c | c  c  c | c  c  c | c  c  c }
	       & \multicolumn{3}{c}{Face}  & \multicolumn{3}{c}{Bird}  & \multicolumn{3}{c}{Car} \\
	       \hline
		   & FID  &  LPIPS   &  KFFA   & FID  &  LPIPS   &  KFFA   & FID  &  LPIPS   &  KFFA \\
		   \hline
           CoModGAN &  8.88   &  0.045   & 52.41   & 11.35 & 0.090 & 52.45 & 6.59 &0.183 & 44.34\\
           CollageGAN & 8.77  &0.069 &66.00 &12.11 & 0.100 & 61.08 &6.57 &0.191 & 48.67 \\
           BAT-Fill & 15.08 & \textbf{0.102} & 75.98 & 37.15& 0.117 & 55.41 & 22.20 & 0.270&51.98\\
           ContrasFill-1 & 8.40 & 0.072 & 74.71 & \textbf{11.29} & 0.151 & 66.06 & \textbf{6.24} & 0.310&63.09 \\
           ContrasFill (Ours) & \textbf{8.36} & 0.075 & \textbf{83.66} & 11.97 & \textbf{0.160} & \textbf{74.58} & 6.46 & \textbf{0.327} & \textbf{82.96} \\ 
           \end{tabular}
	    \caption{ Our method has comparable image quality with the state-of-the-art, but with more diversity.}
	    \label{table:baseline_fid_lpips_kffa}
	\end{center}
\end{table}

\subsection{Image quality and diversity}
Besides our model ContrasFill, we also evaluate one variant of our approach, where we only have one latent space using contrastive learning without explicit latent disentanglement (denoted as ``ContrasFill-1'', see supp for details about this variant). This entangled latent space models all factors together for generation. It is used to show the effectiveness of the contrastive loss on diversity.

Table~\ref{table:baseline_fid_lpips_kffa} shows comparison in terms of image quality (FID) and diversity (LPIPS for overall, KFFA for known factor). Overall, our model has comparable image quality with the state-of-the-art methods, and it performs favorably against all baselines in terms of known factor diversity, especially compared with CoModGAN~\cite{Zhao2021LargeSI} and CollageGAN~\cite{Li2021CollagingCG}. We also have the highest overall diversity on bird and car datasets. Although, BAT-fill~\cite{Yu2021DiverseII} has better LPIPS distance in face dataset, but their image quality is worse (Figure~\ref{fig:baseline} ) and they sometimes generate artifacts, which can often results in larger LPIPS difference. Our model also has better diversity, especially on the known factors, compared with our single code variant (ContrasFill-1).  

We also compared with UCTGAN~\cite{Zhao2020UCTGANDI} which is designed for diverse hole filling. Due to code unavailablity, we only compare with it in CelebA-HQ dataset and use the same setting as theirs. Here we measure diversity LPIPS for full output and only mask region. We grab their numbers and notations (ours first): $I_{out}$: \textbf{0.036} vs 0.030; $I_{out(m)}$: \textbf{0.101} vs 0.092.

\subsection{Disentanglement study}

\begin{figure}[t!]
    \centering
    \includegraphics[width=0.95\textwidth]{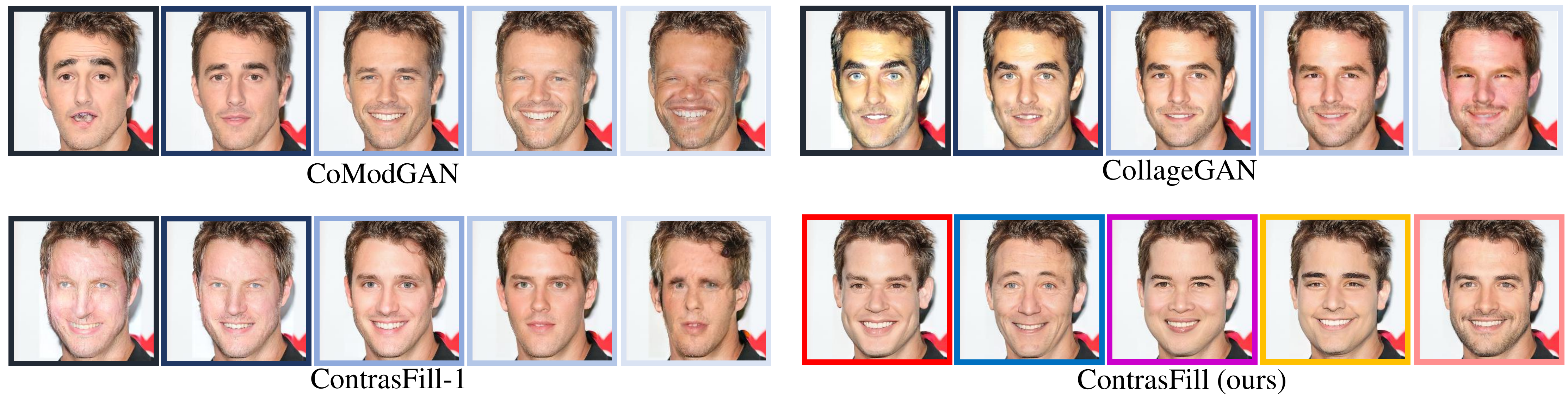}
    \caption{Moving along the discovered identity direction causes changes in non-identity factors such as facial expressions and lighting for baselines. Our results only vary in identity.}
    \label{fig:id_direction}
\end{figure}

We compare with baselines to show that having two explicit latent spaces improves the disentanglement. Recent works~\cite{Voynov2020UnsupervisedDO,harkonen2020ganspace,Shen2020ClosedFormFO} show that postprocessing can be applied to find disentangled latent directions in a pretrained GAN space. Thus we use a supervised method~\cite{Shen2020InterpretingTL} to find known factor directions for CoModGAN, CollageGAN and ContrasFill-1. 
We use pretrained known factor classifiers to get labels for sampled latent codes and then train a linear regressor to find latent directions~\cite{Shen2020InterpretingTL}. We do not compare with BAT-fill since they lack controllability.

Next, we generate images with different known factors. For baselines, given a masked image, we first randomly sample a latent code, and then move along the discovered known direction to generate 10 different results. For our approach, since we have a disentangled space, thus we directly sample 10 different codes in known factor space by fixing unknown code. We calculate KFFA for 1,000 different contexts and report the average number in the Table~\ref{table:disentanglement}. Our model has the best KFFA scores. This demonstrates the benefit of having an explicitly disentangled latent space.

Since we has two spaces, we also conduct an experiment where we fix our known code and randomly sample 10 unknown codes. The last row of Table~\ref{table:disentanglement} shows the average KFFA numbers over 1,000 context images which indicates when unknown factor varies, our known factor is less influenced. Please refer to Figure~\ref{fig:disentanglement} for qualitative results. 

We also visually examine the baselines in Figure~\ref{fig:id_direction}. By moving along the discovered known (identity) direction, those baselines not only change identity to some extent, but also alter other attributes, such as smile, skin tone, and gaze, whereas our sampled results maintain the attributes controlled by the unknown factor while the known factor changes. This means that certain factors such as human identity can not be easily disentangled during the latent direction discovery stage even with explicit supervision. We also show that this is true for vanilla unconditional StyleGAN~\cite{karras-cvpr2019,Karras2019stylegan2} in the supp. Note, we move large steps in two directions on purpose (leftmost and rightmost) to show the full effect of the discovered directions. Details about this study can be found in the supp.

\begin{table}[t!]
    \begin{center}
		\begin{tabular}{ l | c  c  c }
			 & face & bird & car \\
			\hline
			
			CoModGAN     &56.00     &47.55          &44.25 \\
			
			CollageGAN   &57.78     &58.15          &47.49\\

			ContrasFill-1        &67.40     &61.14          &52.01\\
			
			ContrasFill         &\textbf{82.03}     &\textbf{75.20}          &\textbf{83.71}\\
			\hline
			ContrasFill (known fixed)       &26.15     & 21.69
& 35.47 \\
			\hline
		\end{tabular}
	    \caption{ High KFFA shows our latent space is more diverse compared with discovered latent directions in baselines. The last row is a different setting, please refer the text.}
	    \label{table:disentanglement}
	\end{center}
\end{table}

\subsection{Ablation}

\begin{figure}[t!]
    \centering
    \includegraphics[width=0.90\textwidth]{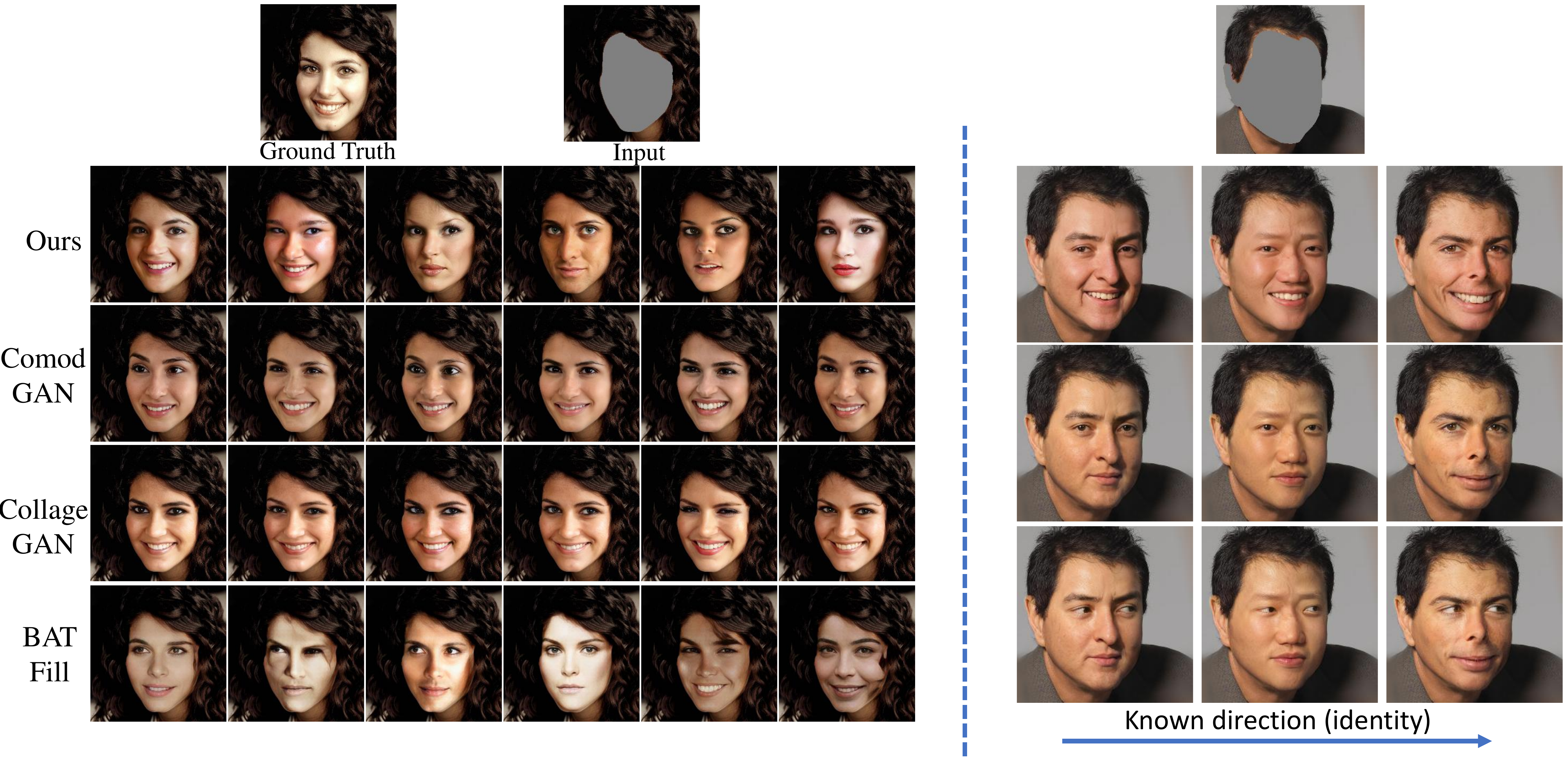}
    \caption{Our model can generate diverse disentangled results in semantic mask case.}
    \label{fig:mask_case}
\end{figure}


\begin{table}[t]
\parbox{.45\linewidth}{
\centering
\begin{tabular}{ c|c|c|c|c } 
& CoMod  & Collage & BAT & Ours \\
\hline
FID   & \textbf{5.73}  & 6.07  &  11.97           & 5.95  \\
LPIPS & 0.029          & 0.029 &  \textbf{0.050}  & 0.048  \\
KFFA  & 51.19          & 58.73 & 72.48            & \textbf{83.39} 
\end{tabular}
}
\hfill
\parbox{.5\linewidth}{
\centering
\begin{tabular}{ c|c|c|c|c|c } 
& concat  & u-k & k-u & repredict & Ours \\
\hline
FID   & 8.64  & \textbf{8.41}  &  8.42  & 14.41 & \textbf{8.41}  \\
LPIPS & 0.048 & 0.052 &  0.056 & \textbf{0.120} & 0.075  \\
KFFA  & 48.99 & 78.19 &  77.58 & \textbf{86.87} & 83.06
\end{tabular}}
\caption{\textbf{(Left) Comparison in mask case.} Our method has comparable image quality but with more diversity on faces when using face masks as inpainting regions. \textbf{(Right) Ablation.} results indicate the effectiveness of bi-modulation (the first three columns) and contrastive loss (4th column).}
\label{table:merge_table}
\end{table}

\noindent\textbf{Other type of masks.} We also analyze our model's performance on the inpainting task where the input mask is of the shape of an object instead of a box. In this case, our model can no longer change the object shape and pose but can still generate diverse appearance in the mask.  Figure \ref{fig:mask_case} (left) shows that our results are more diverse than CoModGAN and CollageGAN, especially on human identity. BAT-fill has worse image quality. We can also achieve disentanglement (Figure \ref{fig:mask_case} right). We compare image quality and diversity (LPIPS for overall, KFFA for known factor), and report numbers in Table~\ref{table:merge_table} (left) for the face dataset. We also study the case of arbitrarily-shaped masks that cover part of the object in random places (e.g., half of face is hidden); see supp for details.

\noindent\textbf{Latent codes injection method.} To study the effectiveness of our bi-modulated convolutions (Figure~\ref{fig:archi}), we try the following alternative approaches: (1) We concatenate $s$ with $t$ and pass the result to fully-connected layers to output a single scale vector to modulate the convolutions (denoted as ``concat''); (2) We use each code to modulate a different set of convolution kernels. And the two sets of modulated convolutions are sequentially applied to image features. Depending on the order, we denote them as ``k-u'' and ``u-k'' (k and u stand for known and unknown codes). The first three columns of Table~\ref{table:merge_table} (right) indicate that, these alternative designs achieve similar image quality, but lower level of diversity. Our bi-modulation is a more direct way to inject information to the generator which makes the learning process easier compared with the ``concat'' alternative. The approach of applying two separate sets of convolutions results in poor diversity. We hypothesize that the later convolution set may undo what is learnt by the previous set as their objective functions are different (factors in two spaces should learn different things).           

\noindent\textbf{The loss choice.} We use the contrastive loss to encourage diversity by forcing images with the same latent codes to have similar factors. Another way to learn this correspondence is to repredict the input code from the resulting image. For example, by sampling a code in the identity space, one can use ArcFace to repredict this code from the generated image. 
Table~\ref{table:merge_table} (right 4th column) shows that replacing the contrastive loss with reprediction loss encourages more diversity, but at the cost of image quality. This is because a foreground generation model needs to consider the compatibility between the sampled latent code and the input context. For example, if the context contains light skin pixels on the neck, then latent codes that generate dark-skinned faces are not compatible.  
However, the reprediction loss forces the model to synthesize a dark-skinned face, which may not look real according to the discriminator. However if a contrastive loss is applied, which considers the \emph{relative distance} in the feature space, then the model can adjust the input identity code based on the context information to synthesize a face that looks more plausible in the context.

\section{Conclusion and Limitations}
We propose ContrasFill, a novel approach for diverse and controllable foreground generation by contrastive learning. We demonstrate superior diversity and controllability over previous work. Our method has some limitations. We found that our model is sometimes sensitive to the pretrained classifier which may be biased due to the training data. For example, certain types of car are more common in certain color (e.g., van are usually white). Thus our model may also be biased. 

\noindent\textbf{Acknowledgement} This work was supported in part by Sony Focused Research Award, NSF CAREER IIS-2150012, Wisconsin Alumni Research Foundation, and NASA 80NSSC21K0295.

\clearpage
\bibliographystyle{splncs04}
\bibliography{egbib}
\clearpage
\section*{Appendix}
In this supplementary material, we first introduce our model architecture and implementation details in Section~\ref{sec:architecture_details}, followed by the dataset details in Section~\ref{sec:datasets_details}. In Section~\ref{sec:visual_examples_of_the_approach}, we provide a visual example to better understand our approach in the main paper. Next, we give details about disentanglement study conducted in the main paper and also show this study on vanilla StyleGAN in Section~\ref{sec:identity_direction_study}. Finally, we show more studies and qualitative results in Section~\ref{sec:additional_qualitative_results_and_studies}.

\section{Architecture and implementation details}
\label{sec:architecture_details}

\begin{figure}[t!]
    \centering
    \includegraphics[width=0.95\textwidth]{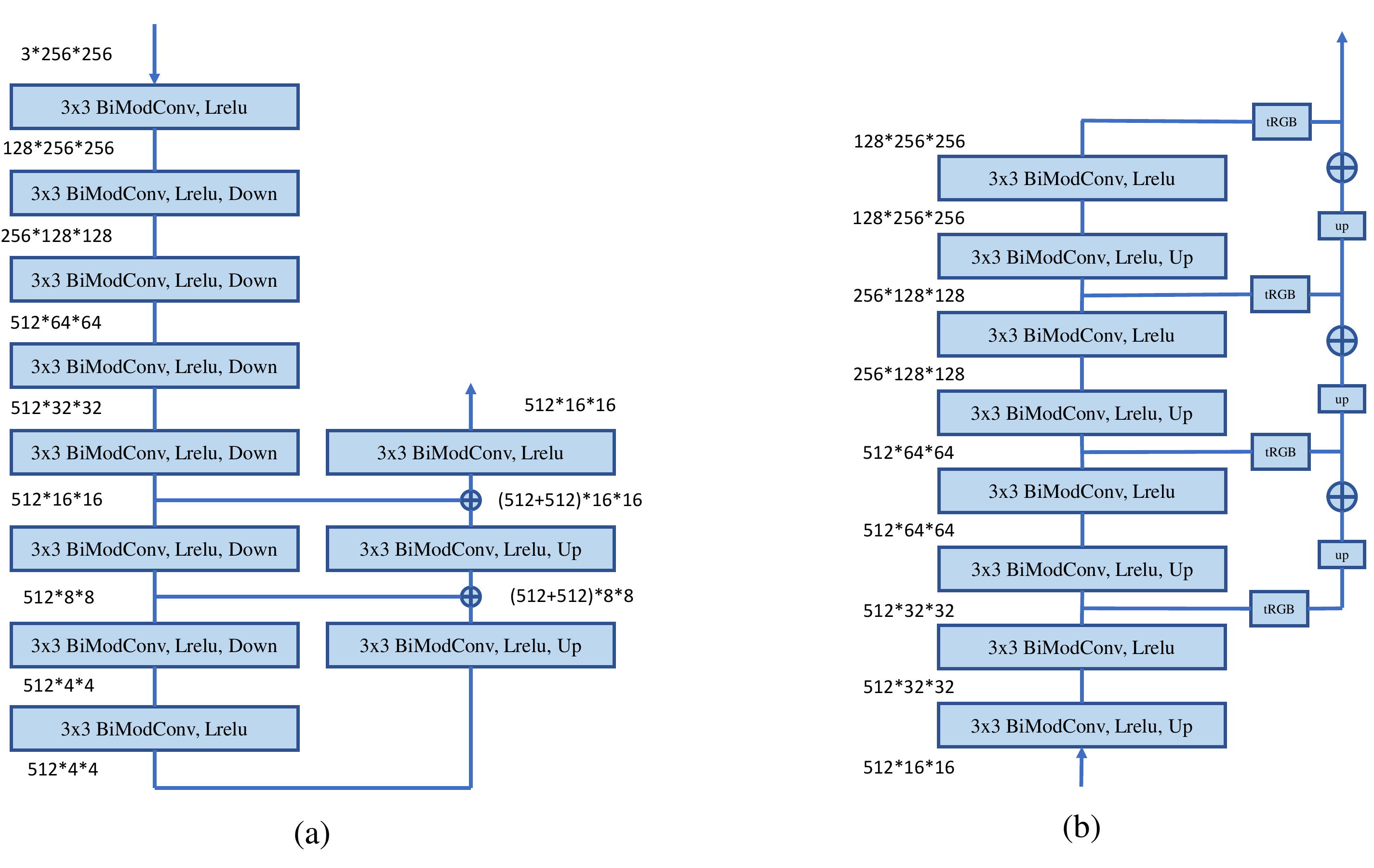}
    \caption{Architecture details.}
    \label{fig:architecture}
\end{figure}

Our model is an encoder and decoder architecture. Figure~\ref{fig:architecture} shows the encoder (left) and decoder (right) details. The input is the masked image of size $256 \times 256$. It consists of a bunch of Bi-modulated convolution and leaky relu activation functions. Inspired by~\cite{Li2021CollagingCG} to preserve better spatial alignment information with the input inpainting mask, once the feature reaches to $4 \times 4$ we upsample it to $16 \times 16$ with lateral connections. In decoder, we have two bi-modulated convolution for each resolution and one $to\_rgb$ layer as output skip connection since it is beneficial for gradient update~\cite{karras-cvpr2019,Karras2019stylegan2}. Note we do not show two input codes for bi-modulated convolution block in both encoder and decoder for simplicity. Please refer Figure 3 in the main paper for details about bi-modulation. 

We train our model for 200k iterations with learning rate of 0.002 using Adam optimizer. We set the weight of known factor loss $\lambda_1$ as 1, and weight of unknown factor loss $\lambda_2$ as 0.1 (for face and bird) and 5 (for car) for the initial 100k iterations. We decrease both $\lambda_1$ and $\lambda_2$ by 10 times for the next 100k iterations.

\section{Dataset details}
\label{sec:datasets_details}

For face dataset, we use the official face parsing annotations from CelebA-HQ\cite{lee-cvpr2020} to define our face region. Specifically, the following officially defined semantic regions are merged: 'skin', 'nose', 'eye\_glass', 'l\_eye', 'r\_eye', 'l\_eyebrow', 'r\_eyebrow', 'l\_ear', 'r\_ear', 'mouth', 'u\_lip', 'l\_lip'. The union of these region is defined as face mask in all our experiments. We convert mask to box when we mask out inpatining region in the main paper. For the extra training data from FFHQ~\cite{karras-cvpr2019}, we do the same process. 

For bird and car which are obtained from the LSUN dataset~\cite{Yu2015LSUNCO}, we first only choose images whose both sides are greater than 400 resolution and at least one side is greater than 512 resolution. Then we use a pretrained MaskRCNN~\cite{He2020MaskR} from Detectron2~\cite{wu2019detectron2} to select images. Specifically, the most confident detected object needs to be our target object ('bird' for our bird dataset, 'truck' or 'car' for our car dataset) and the most confident object also needs to be the biggest mask in its image.  

We use the pretrained ArcFace~\cite{Deng2019ArcFaceAA} as the known factor encoder for face dataset, and train a classifier~\cite{du2020fine} on CUB dataset for bird known factor encoder. For car, we first group images in the Stanford car dataset~\cite{KrauseStarkDengFei-Fei_3DRR2013} into 8 classes based on their shape and then train a classifier~\cite{du2020fine}. We choose the following shapes: 'SUV', 'Sedan', 'Hatchback', 'Coupe', 'Convertible', 'Wagon', 'Cab', 'Van' from their annotation names.

\section{Visual explanations of our approach}
\label{sec:visual_examples_of_the_approach}
\begin{figure}[t!]
    \centering
    \includegraphics[width=0.65\textwidth]{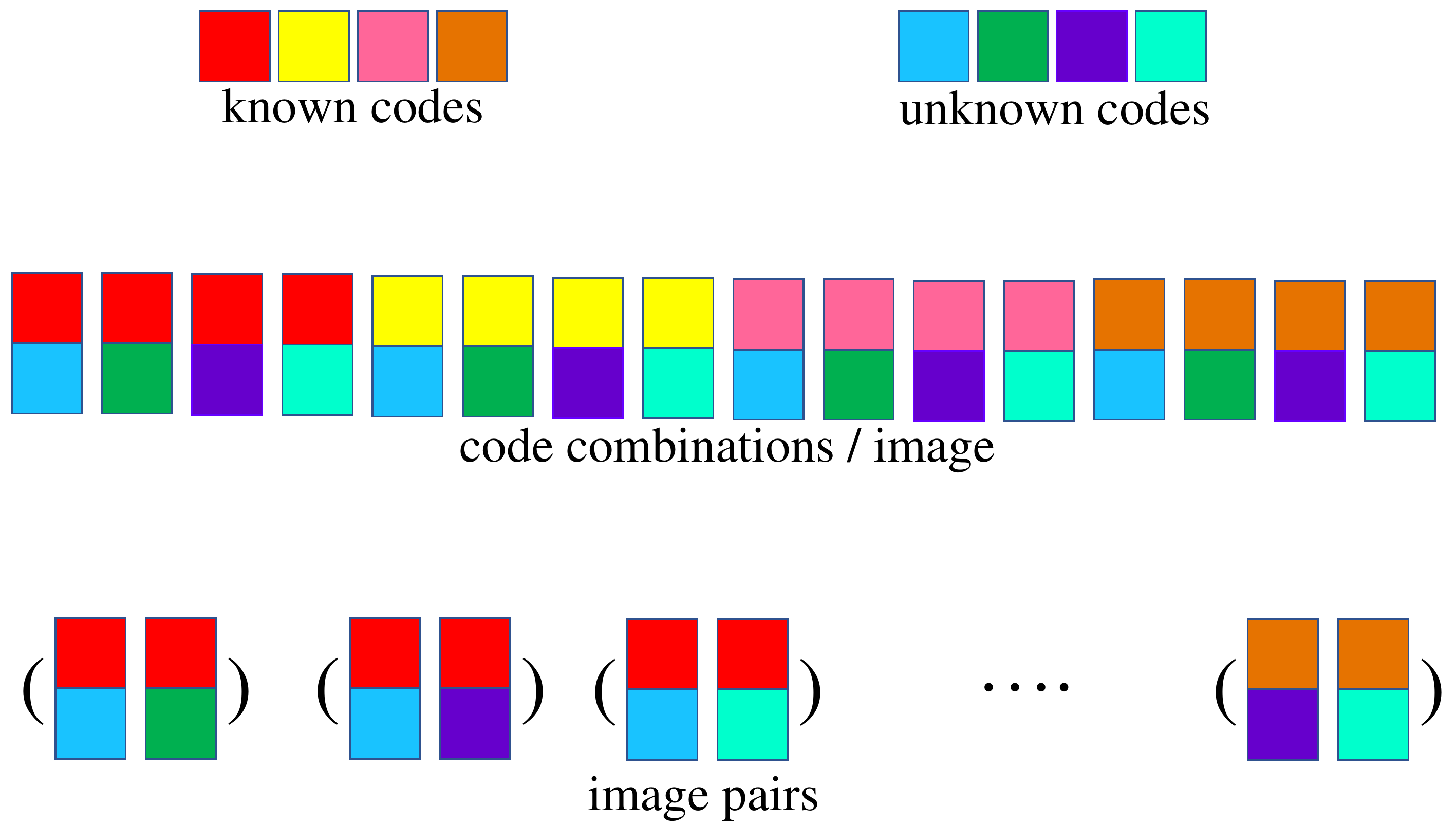}
    \caption{4 warm/cold colors represent known/unknown codes. Totally, there are 16 code combinations in the middle. Bottoms shows any two different combinations can be form as an image pair (Eq1 in the main paper). }
    \label{fig:codes_combinations_pairs}
\end{figure}

\begin{figure}[t!]
    \centering
    \includegraphics[width=0.65\textwidth]{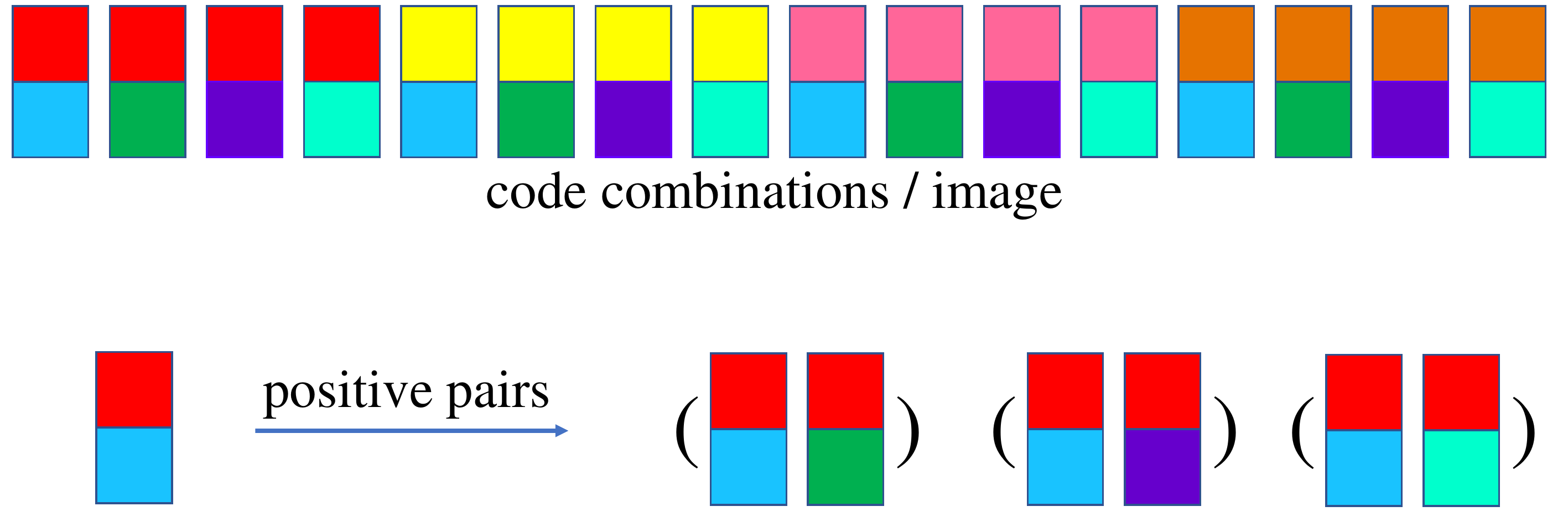}
    \caption{This is an example for the notation $P_{ \textbf{k},u}$: among 16 code combinations (or image), for the given image (red code + blue code), we have three positive pairs for it.}
    \label{fig:positive_pairs}
\end{figure}

\begin{figure}[t!]
    \centering
    \includegraphics[width=0.65\textwidth]{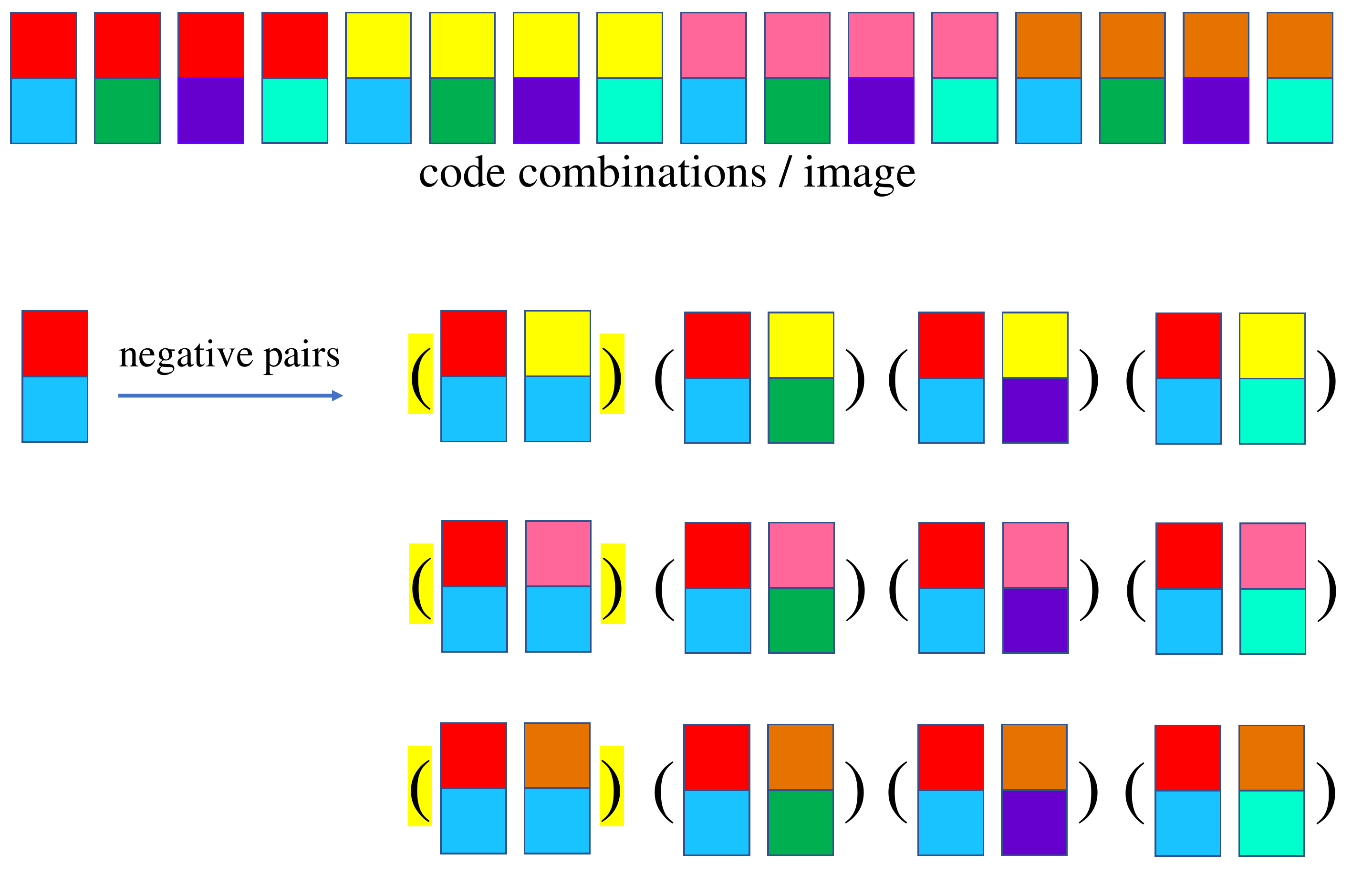}
    \caption{This is an example for the notation $N_{ \textbf{k},u}$: among 16 code combinations (or image), for the given image (red code + blue code), we have 12 negative pairs for it. Parentheses of hard negative pairs are highlighted.}
    \label{fig:negative_pairs}
\end{figure}

\begin{figure}[t!]
    \centering
    \includegraphics[width=0.65\textwidth]{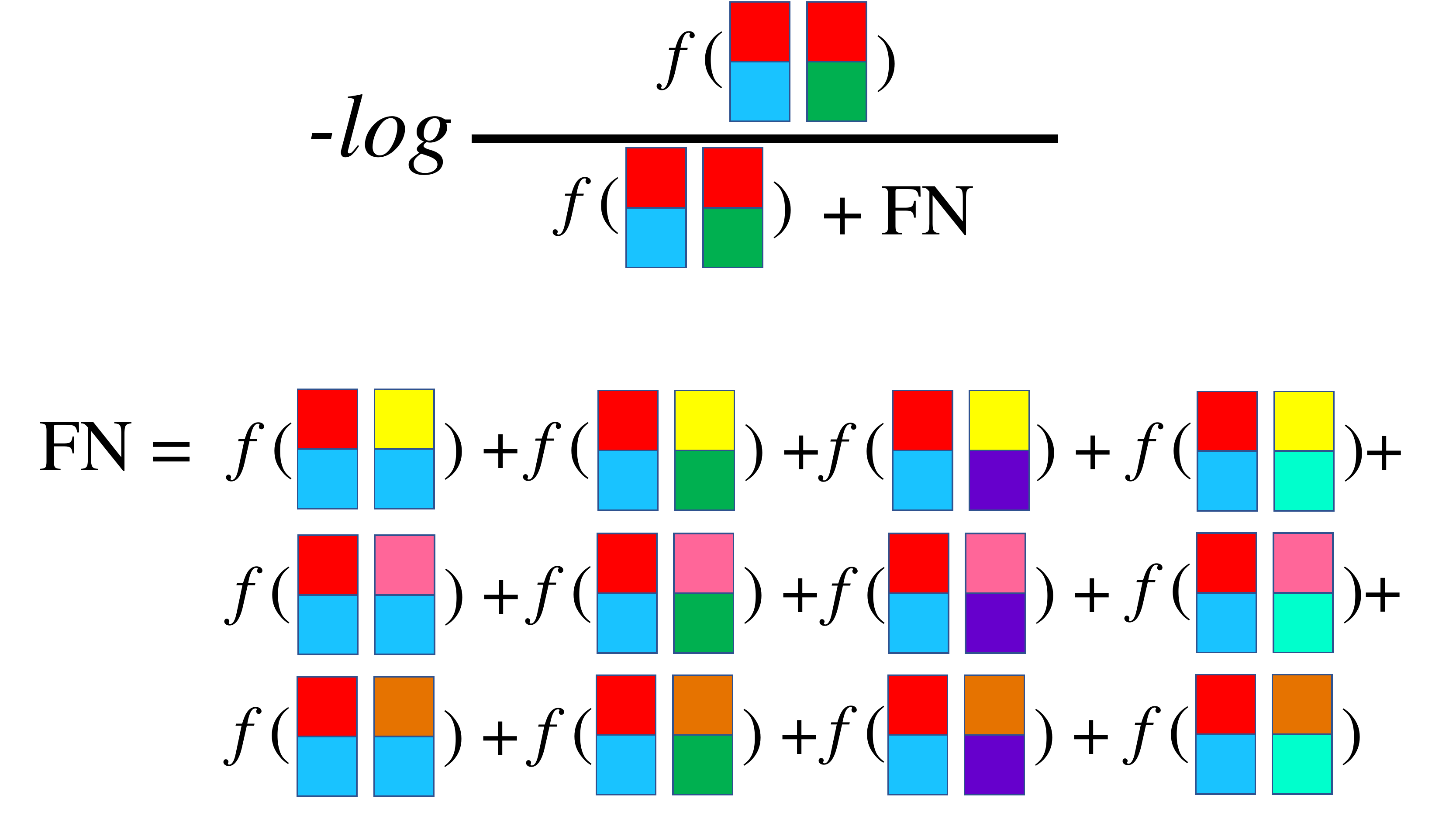}
    \caption{Visual example of one case of Eq3 in the main paper.}
    \label{fig:loss}
\end{figure}

To understand our approach more clearly, we use a visual example to demonstrate key notations and equations defined in the main paper. Note Figure~\ref{fig:codes_combinations_pairs}  to Figure~\ref{fig:loss} are describing different terms in the same example. 

Suppose we only have 4 known codes (warm colors: red yellow, pink and orange) and 4 unknowns codes (cold colors: blue green, purple and cyan) as shown in Figure~\ref{fig:codes_combinations_pairs}. Like mentioned in the main paper, we ignore the input context image $I$ as it is irrelevant to our contrastive learning, then each combination (in the middle of Figure~\ref{fig:codes_combinations_pairs}) represents one resulting image $S$. And each parenthesis at the bottom means image pair (Eq1 in the main paper). In this case, there are $\binom{16}{2}$ image pairs in total. 

Again we only consider the case in the known space for simplicity. The positive pair refers to two images sharing the same known code (e.g., top part color in the code combination should be same). In the main paper, we define $P_{ \textbf{k},u}$ as a set of all positive pairs \emph{associated} with the code combination $(k,u)$ in the known space. Here is how to understand our notation: $P$ means positiveness, $(k,u)$ indicates which code combination (or image), the bold letter defines we are considering from known space perspective. In Figure~\ref{fig:positive_pairs}, we show one example of $P_{ \textbf{k},u}$ for the image whose known code is red and unknown code is blue. 

In the main paper, $P_{ \bm{K} }$ is defined as \emph{all} the positive pairs in the known space. In this visual example, there are $4*\binom{4}{2}$ pairs in total. Here is how to break it down: consider only one color in the known space, say red, and there are four images with red code. Among these four, we can choose $\binom{4}{2}$ pairs that are positive. With the same strategy applied to the other three colors, $| P_{ \bm{K} } |$ is $4*\binom{4}{2}$ in this case. To emphasize their differences: the $P_{ \textbf{k},u}$ is \emph{associated} with one particular code combinations (or image) we are considering, whereas $P_{ \bm{K} }$ consists of \emph{all} positive pairs in the known space.

For the negative pair, we consider two images not sharing the same known code (top part color is different). Similarly, we also define $N_{ \textbf{k},u}$ as a set of all negative pairs \emph{associated} with the code combination $(k,u)$ in the known space. Figure~\ref{fig:negative_pairs} demonstrate 12 negative pairs for the image with red known code and blue unknown code. Among those 12 negative pairs, there are 3 hard negative pairs (parentheses are highlighted) which has the same unknown codes. Since in the hard negative pairs, features in the unknown space is the same, thus the network has to synthesize different factors according to the different known codes such that they can be recognized differently by the known factor encoder (orange one in the main paper Figure 2).

We show one example of loss function defined in the main paper Eq3. The Eq3 in the main paper is defined for \emph{one} positive pair. Suppose we consider the image (red+blue) and image (red+green), the loss function calculated for this pair is visually shown in Figure~\ref{fig:loss}. The $f( \cdot, \cdot )$ means the similarity between two images defined as in Eq2 in the main paper. Conceptually, we push the numerator high and $FN$ in the denominator low. Since we have $4*\binom{16}{2}$ positive pairs in this visual example, the final known space loss (Eq4 in the main paper) is averaged across all positive pairs. Situation in the unknown space is the same which only requires viewing codes from a different perspective. 

\begin{figure}[t!]
    \centering
    \includegraphics[width=0.65\textwidth]{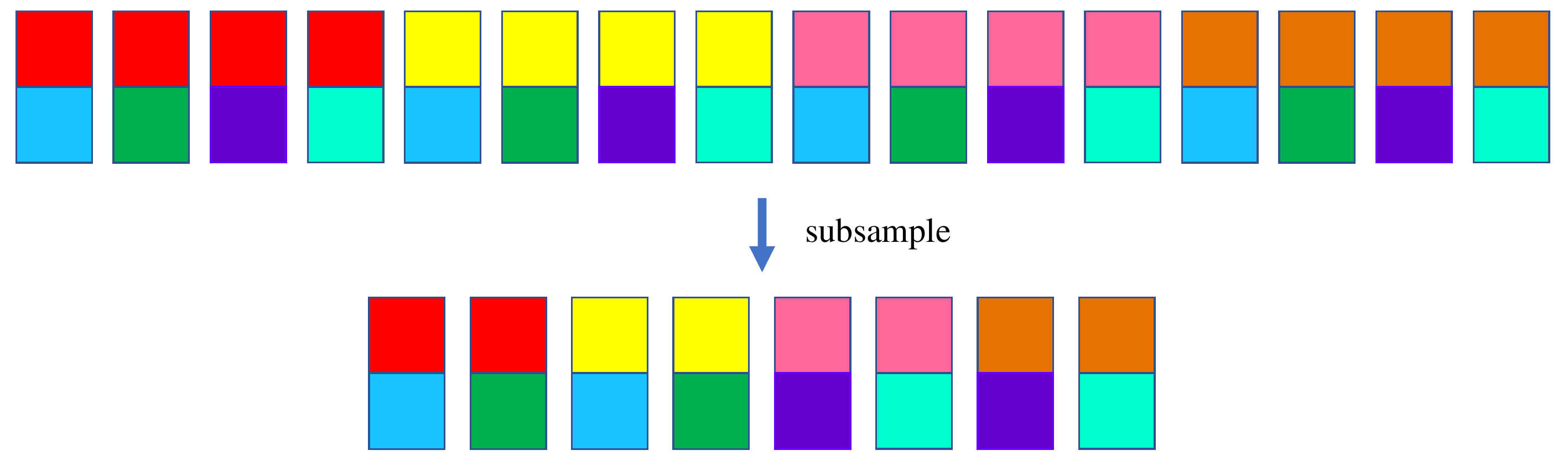}
    \caption{Subsample 8 code combinations from 16.}
    \label{fig:subsample}
\end{figure}

As mentioned in the main paper, in practice, we subsample code combinations (or images) such that each image has one hard negative pair in both spaces due to GPU memory constraint. Figure~\ref{fig:subsample} shows the subsample result for our visual example. In practice, to increase code diversity in each space, we use 8 known and 8 unknown codes in our implementation, which will have 64 code combinations (or images) in total. We use the same subsample strategy as shown in Figure~\ref{fig:subsample} resulting batch size of 16 during training. 

In the main paper, we also evaluate one variant of our approach: ContrasFill-1 which only has unknown code. We apply contrastive loss for this model to increase the general diversity. This variant can be served as a baseline, which can be used to evaluate the effectiveness of contrastive loss on the diversity. And our final model, ContrasFill can be used to compare with it to study the benefit of two explicit disentangled spaces. To train such variant model, we only need to: (1) sample one set of code from unknown space each time (images with the same/different codes defined as positive/negative pairs), and apply contrastive loss in this space only. (2) replace all bi-modulated convolution in our model with normal modulated convolution proposed in the StyleGAN2~\cite{Karras2019stylegan2}. It is worth mentioning that although in this case we only have one set of code from unknown space, images with the same code do not mean that they are identical, since we always randomly sample the input image $I$ during training as mentioned in the main paper.

\section{Known factor direction study}
\label{sec:identity_direction_study}

In this section, we first provide more details about section 4.3 of the main paper and then discuss identity direction in unconditional StyleGAN.

\subsection{Details about Section 4.3 in the main paper}
In the main paper Section 4.3, we study the disentanglement of known factor direction in the latent spaces of baselines (CoModGAN, CollageGAN, ContrasFill-1). Here we describe more details. 

Following~\cite{Shen2020InterpretingTL}, we train a linear regressor based on latent codes and their identity labels. Specially, we first randomly sample 100k images using 100k latent codes; and then use the pretrained known factor classifier to get penultimate features of these images. Then we try to group these 100k features into 1k clusters using K-means. For each cluster, we choose top 10 features which are closest to their clustering center. Since each image feature is associated with its latent code, after doing so, we have 10k codes with their group labels. We then train a linear regressor, which predicts a scalar value for each code. The objective for this regressor is to predict same/different value if codes belonging/not belonging to the same cluster. We use the contrastive loss~\cite{Chen2020ASF} to train this regressor. After training, the weight of this regressor should indicate the latent direction for the known factor. For example, in the case of face, if one moves along this discovered direction, then the ideal change in the image should be related with identity.

After discovering the known factor directions for baselines. We randomly sample 1k images and for each image we sample 10 different results for the same mask by moving along the discovered known direction $d$. Since it is hard to define how far one should move along this direction, thus we use the trained discriminator to monitor this process. Specifically, for each image, we first sample 1 result and denote its latent as $w_o$. We obtain its realness score $r$ (between 0 to 1) base on the trained discriminator, and then we define a lower bound $l$ as $r-0.1$. To sample along this direction, we randomly sample a step $s$ from $\mathcal{U}_{[-R,R]}$, where $R$ is a scalar. Then the new image will be synthesized by the latent code $w_o+s*d$. We only choose the image whose realness score is above than lower bound $l$. Otherwise, it indicates we step too far such that the image is not in the distribution anymore. If this happens (e.g., realness is lower than lower bound) we call it a miss. We empirically set $R$ such that the chance of missing is around 10\%, and this is to make sure we do not have a too small step in the direction $d$. 

As the results shown in the main paper, the postprocessing method is not as good as having explicit disentangled latent space even if they all access to the same supervision. 

\subsection{Identity direction in StyleGAN}
The experiment in the main paper section 4.3 shows that directions for certain factors such as human identity can not be easily disentangled in the StyleGAN based inpainting model. For completeness, we find that this is also true for the unconditional StyleGAN. 

Starting with CelebA-HQ dataset where we know identity labels and a pretrained StyleGAN, we first use PSP~\cite{richardson2021encoding}, the state-of-the-art StyleGAN face inversion encoder, to encode these images into the StyleGAN latent space. Then we treat latent codes and their corresponding identity labels as training data. Similarly, following the supervised method~\cite{shen2020interfacegan}, we train a linear regressor to group all latent codes. Once it is trained, its weight should indicate the direction for identity in the vanilla StyleGAN latent space.

\begin{figure}[t!]
    \centering
    \includegraphics[width=0.80\textwidth]{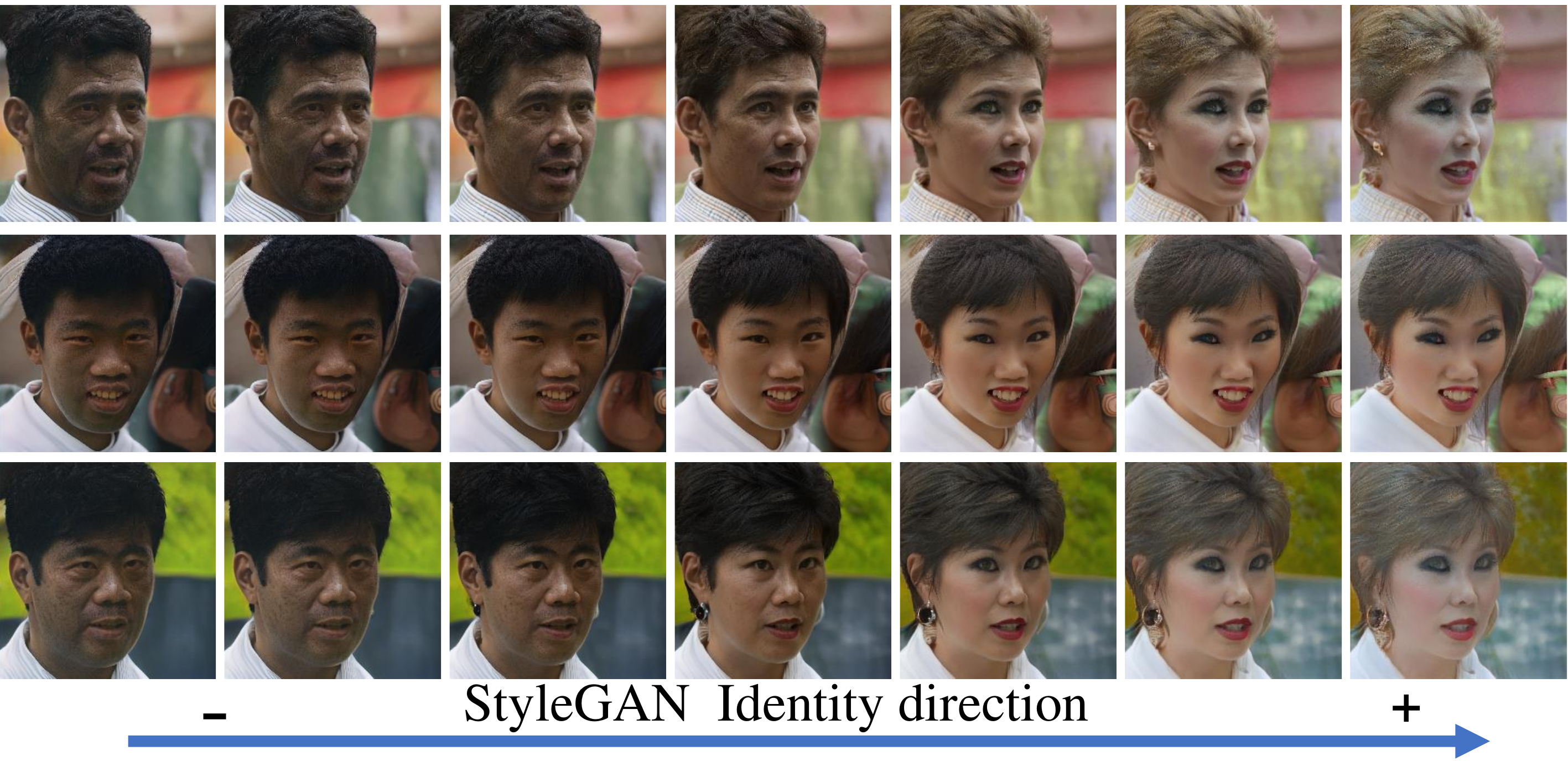}
    \caption{Moving along the discovered identity direction in the StyleGAN latent space.}
    \label{fig:stylegan_id}
\end{figure}

Figure~\ref{fig:stylegan_id} shows the results where we move along the discovered identity direction for three samples (middle one in each row). This direction is able to change identity, however, the resulting images still resemble images in the middle. Also certain attributes change as well such as eyes open more and makeup become heavier when moving to the right. This shows that the identity direction can not be easily disentangled even in the unconditional StyleGAN.

\begin{figure}[t!]
    \centering
    \includegraphics[width=0.6\textwidth]{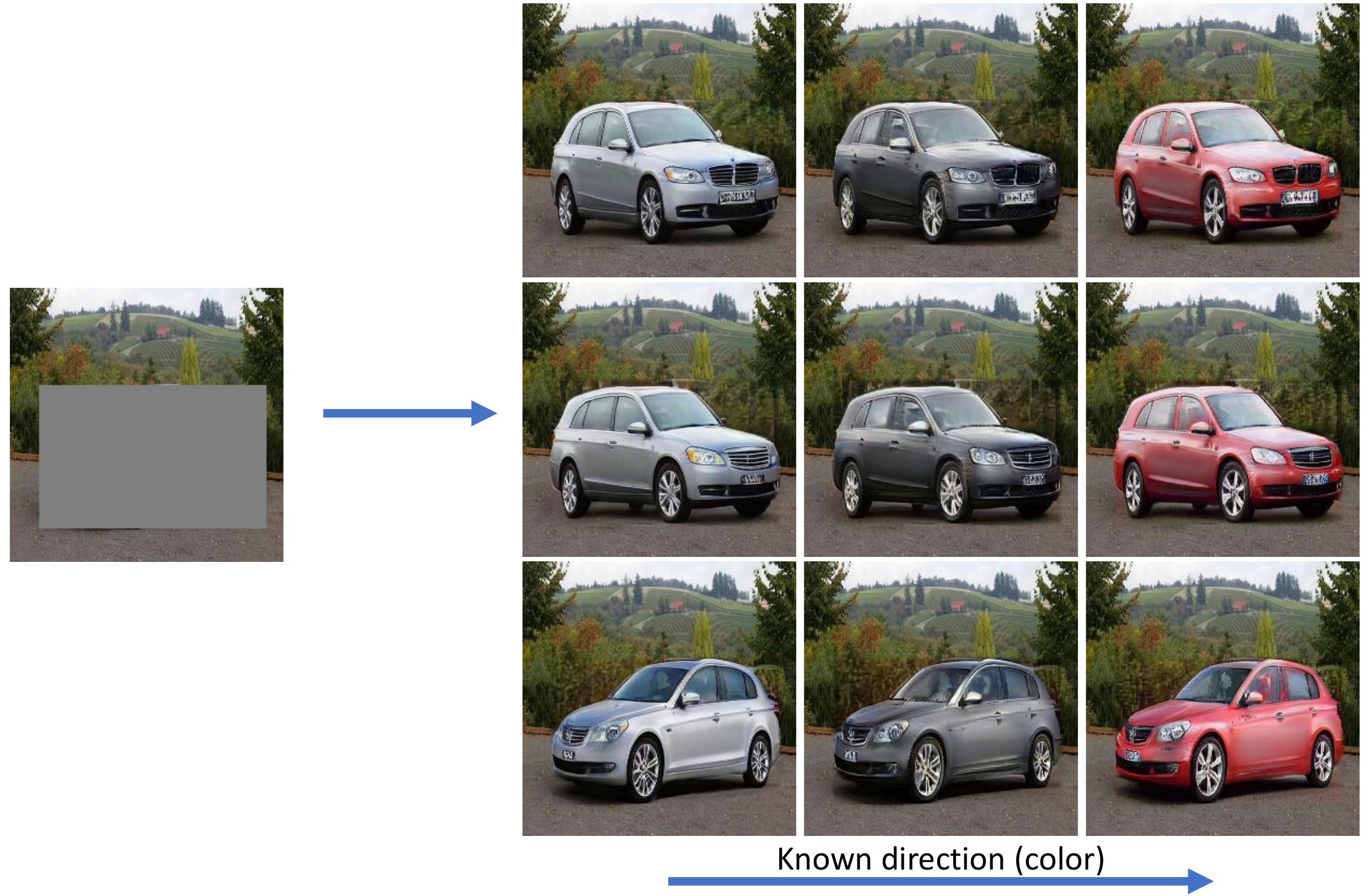}
    \caption{Each column has the same known factor (color), and each column has the same unknown factor which controls the object shape and pose.}
    \label{fig:color_known}
    \vspace{-3pt}
\end{figure}

\section{Additional qualitative results and studies}
\label{sec:additional_qualitative_results_and_studies}

\begin{figure}[t!]
    \centering
    \includegraphics[width=0.80\textwidth]{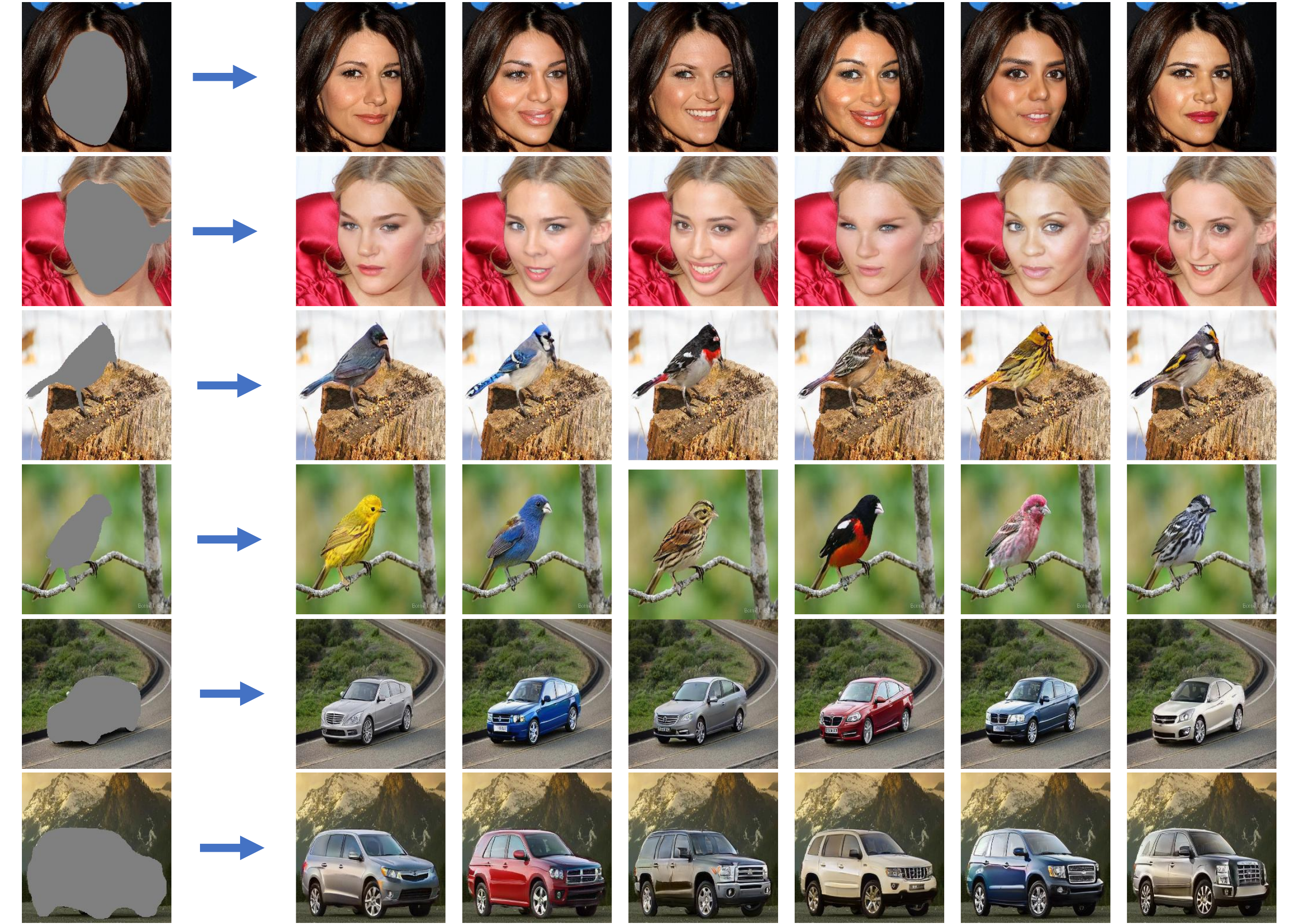}
    \caption{We can achieve diverse samples when filling region is object semantic mask.}
    \label{fig:mask_diversity}
    \vspace{-3pt}
\end{figure}

\begin{figure}[t!]
    \centering
    \includegraphics[width=0.80\textwidth]{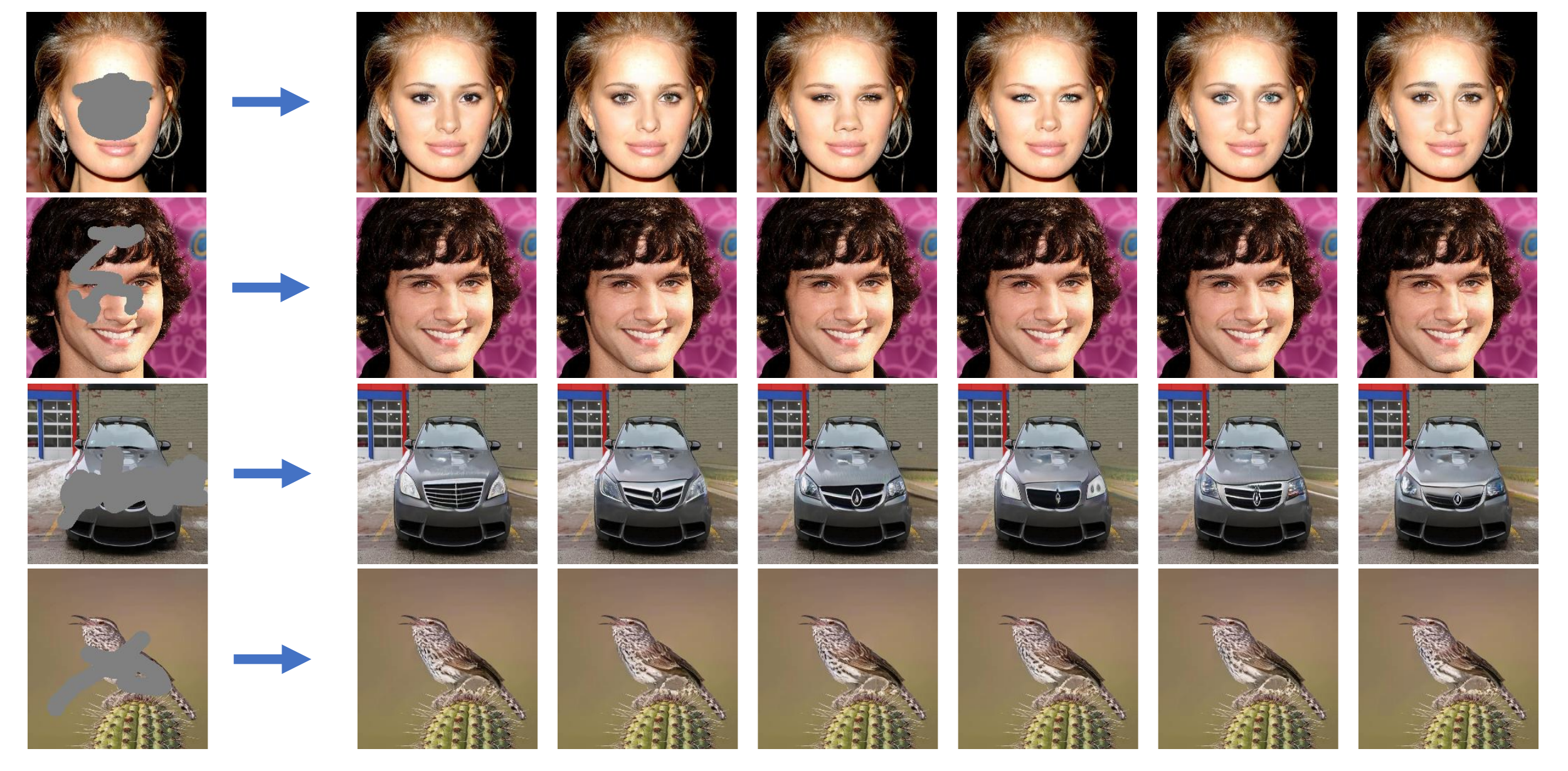}
    \caption{If masked region allows diversity (row 1, 3), our model still can generate diverse results. Our model generate almost deterministic results when partially masked region is strongly correlated with the context regions (row,2,4)}
    \label{fig:random_mask}
    \vspace{-3pt}
\end{figure}

Here we provide additional qualitative and quantitative studies about our method. 

\noindent\textbf{Generalization across different known factors.} To demonstrate the generalization of our approach on the known factor, we train a car color classifier as our known factor feature extractor. We use the Stanford car dataset~\cite{KrauseStarkDengFei-Fei_3DRR2013} as training data, and average pixels of foreground object in each image and run k-means to group all averaged colors into 8 classes. Then we train our model on the same car data we have. Figure~\ref{fig:color_known} shows the disentanglement results, where each column has the same known code controlling the color, and each row has the same unknown code which controls object pose and shape. Please compare this result with the car result in Figure 6 of the main paper where the known factor is the car shape.


\noindent\textbf{Other mask regions instead of box.} In the main paper figure 8, we show results where the foreground region is the object semantic mask, instead of box, for face dataset. Here we show more results on all three datasets in Figure~\ref{fig:mask_diversity}. For face, we can still generate diverse identities and expressions. For bird, we can generate different textures. For car, we can generate different color, headlights and car grills. Note that we set known code as color in car dataset as shape can not be changed when the foreground region is object semantic mask. We also quantitatively measure our generation quality (FID), overall diversity (LPIPS) and known factor diversity (KFFA) with other baselines in  three datasets when the missing region is object mask, and we report numbers in Table~\ref{table:mask_baseline_fid_lpips_kffa}. Overall, we have comparable image quality with CoModGAN~\cite{Zhao2021LargeSI} and CollageGAN~\cite{Li2021CollagingCG} as reflected by FID, but we can generate much more diverse results, especially in terms of known factors. Although BAT-Fill has high diversity also, they suffer from image quality as reflected by the FID. 

Our problem setting is foreground object generation where the entire foreground region should be masked, but we also tried to study how does our model behave if object is masked partially. As shown in the Figure~\ref{fig:random_mask}, if there is loose correlation between masked region and context, we can still generate diverse results. E.g., if we masked both eyes (1st row) and car headlights and grills (3rd row), our model can still generate diverse results. But if masked regions have strong correlation with the context and can be inferred only in a fixed way, then our generated results are deterministic with limited variations.  E.g., if we only mask one eye (2nd row) or part of the bird (4th row), then results are almost deterministic with only small variation (such as texture patterns of the birds in the 4th row if check closely). Our model behaviour is actually expected; since if a missing region is heavily dependent on the context, then results should be almost the same. Nevertheless, in our problem setting of foreground generation, we generate diverse results as entire object is masked and missing regions are not strongly correlated with the context.

\begin{table}[t!]
\footnotesize
    \begin{center}
		\begin{tabular}{ c | c  c  c | c  c  c | c  c  c }
	       & \multicolumn{3}{c}{Face}  & \multicolumn{3}{c}{Bird}  & \multicolumn{3}{c}{Car} \\
	       \hline
		   & FID  &  LPIPS   &  KFFA   & FID  &  LPIPS   &  KFFA   & FID  &  LPIPS   &  KFFA \\
		   \hline
           CoModGAN &  \textbf{5.73}   &  0.029   & 51.19   & 7.92 & 0.030 & 27.35 & \textbf{5.77} & 0.146 & 57.53 \\
           
           CollageGAN & 6.07  &  0.029 &   59.73   & 8.99 & 0.023 & 24.70 &6.02 &0.171 & 58.88 \\
           
           BAT-Fill &  11.97 &    \textbf{0.050} & 72.48 & 31.25& 0.080 & 55.36 & 19.88 & 0.200 & 60.44 \\
           
           ContrasFill (Ours) & 5.95 & 0.048 & \textbf{83.39} & \textbf{7.74} & \textbf{0.101} & \textbf{72.23} & 5.95 & \textbf{0.201} &  \textbf{77.86} \\ 
           \end{tabular}
	    \caption{ Our method has comparable image quality with the state-of-the-art, but with more diversity.}
	    \label{table:mask_baseline_fid_lpips_kffa}
	    \vspace{-5pt}
	\end{center}

\end{table}

\begin{figure}[t!]
    \centering
    \includegraphics[width=0.75\textwidth]{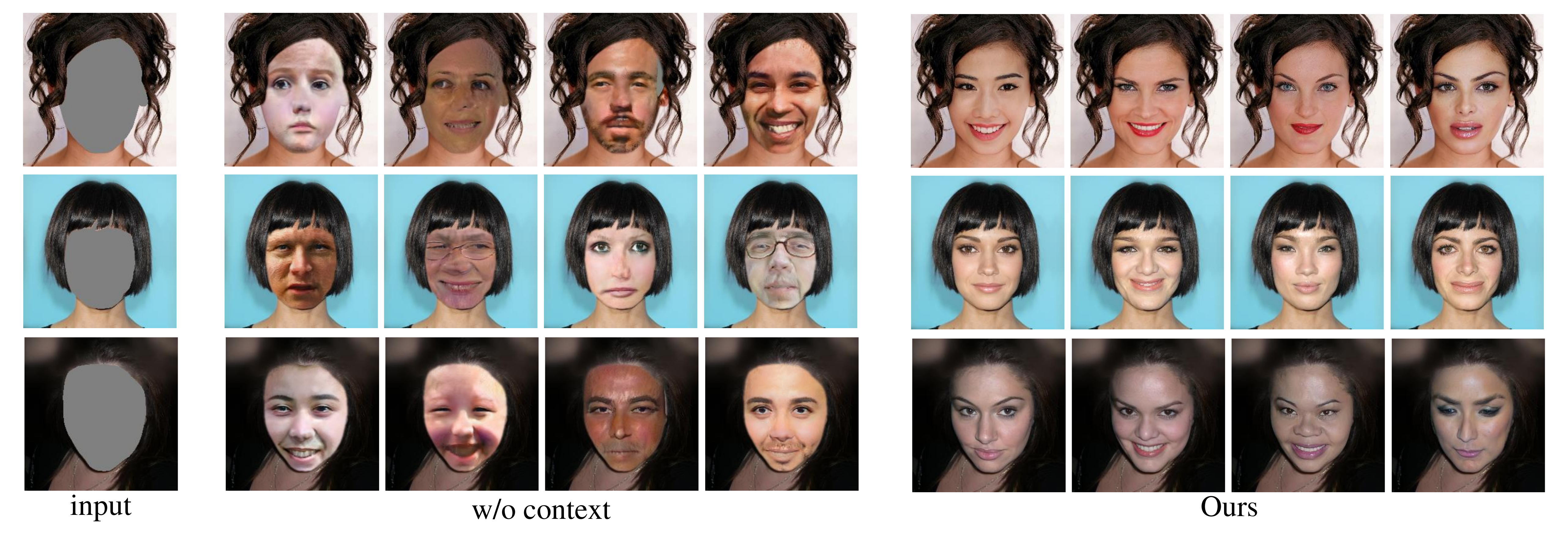}
    \caption{A model will generate inconsistent foreground without considering context.}
    \label{fig:no_context}
    \vspace{-3pt}
\end{figure}

\noindent\textbf{Importance of context.} Although in our setting the entire object is generated,  the model still needs to use context for a correct generation. To study the importance of context, one naive baseline is to treat this as a pure generation problem: generating a foreground object without considering context and then pasting the result back. Thus we design a baseline where the model is only conditioned on the mask to confine the generation region, and after generation we copy and paste the generated result back to the masked context image. As  Figure~\ref{fig:no_context} shows that if we generate a foreground object without considering context information, the generated objects may not be compatible with original surroundings in terms of lighting, pose and other factors.

\begin{figure*}[t!]
    \centering
    \includegraphics[width=0.90\textwidth]{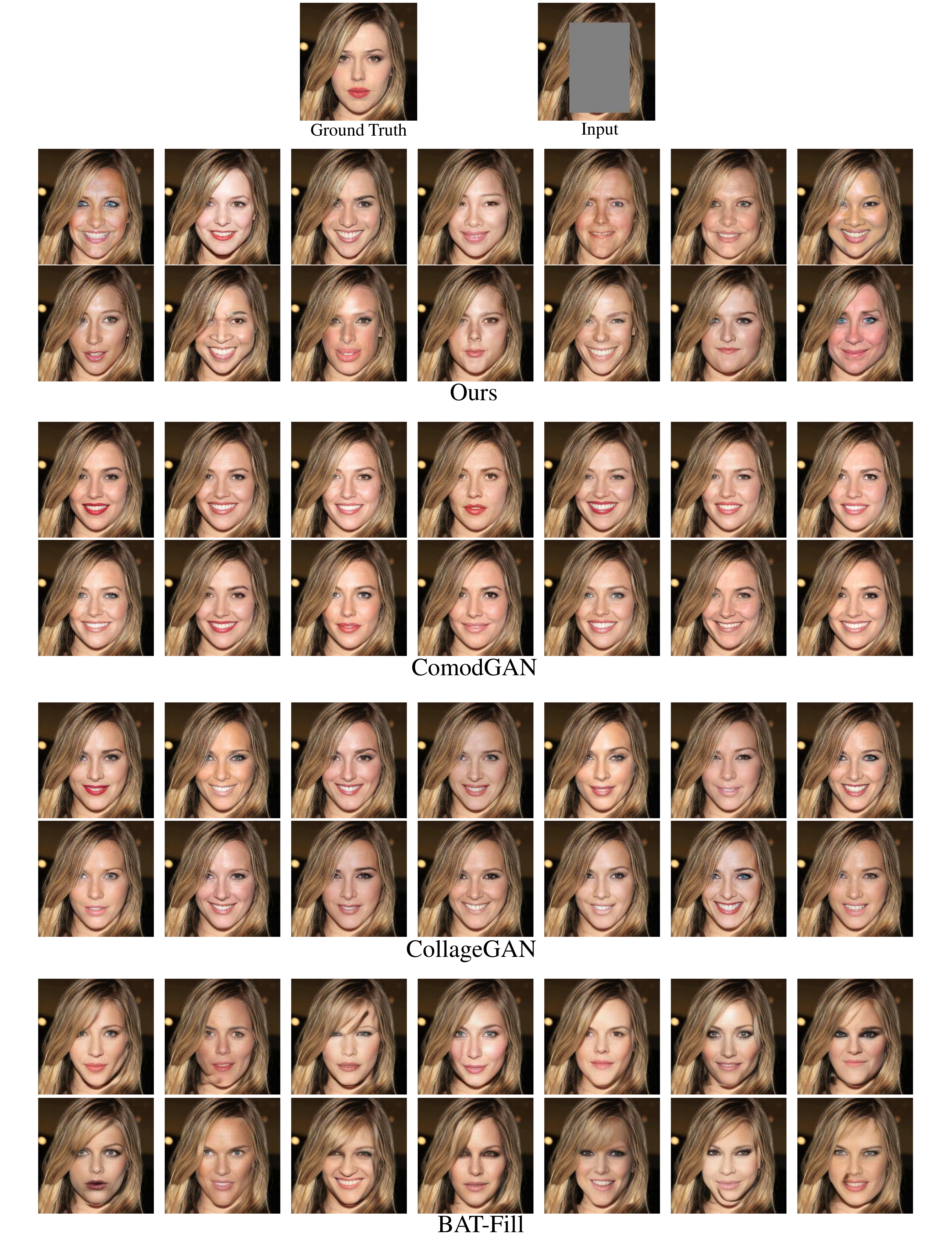}
    \caption{Random samples on the face dataset.}
    \label{fig:baseline_face}
\end{figure*}

\begin{figure*}[t!]
    \centering
    \includegraphics[width=0.90\textwidth]{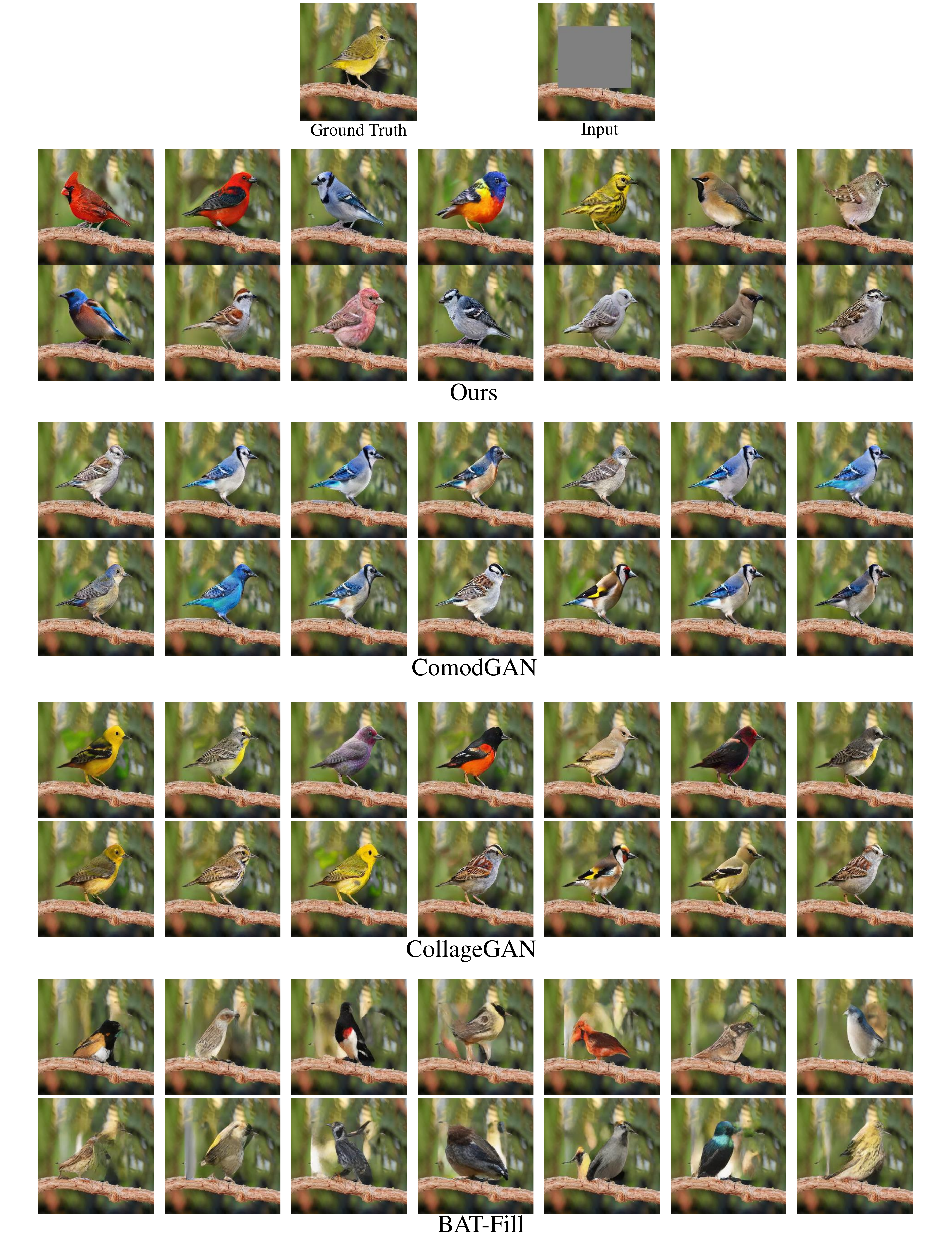}
    \caption{Random samples on the bird dataset.}
    \label{fig:baseline_bird}
\end{figure*}

\begin{figure*}[t!]
    \centering
    \includegraphics[width=0.90\textwidth]{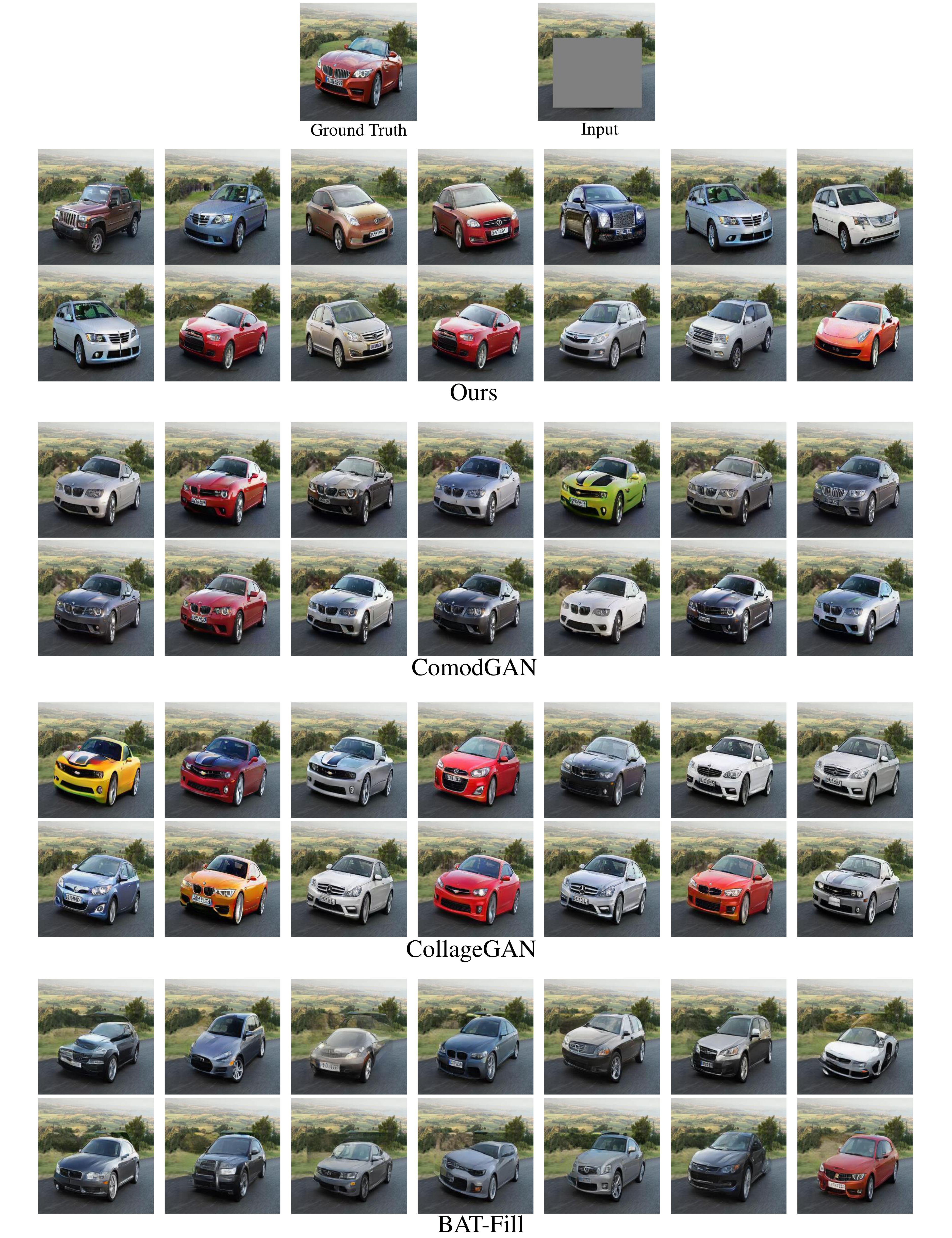}
    \caption{Random samples on the car dataset.}
    \label{fig:baseline_car}
\end{figure*}

\begin{figure*}[t!]
    \centering
    \includegraphics[width=0.98\textwidth]{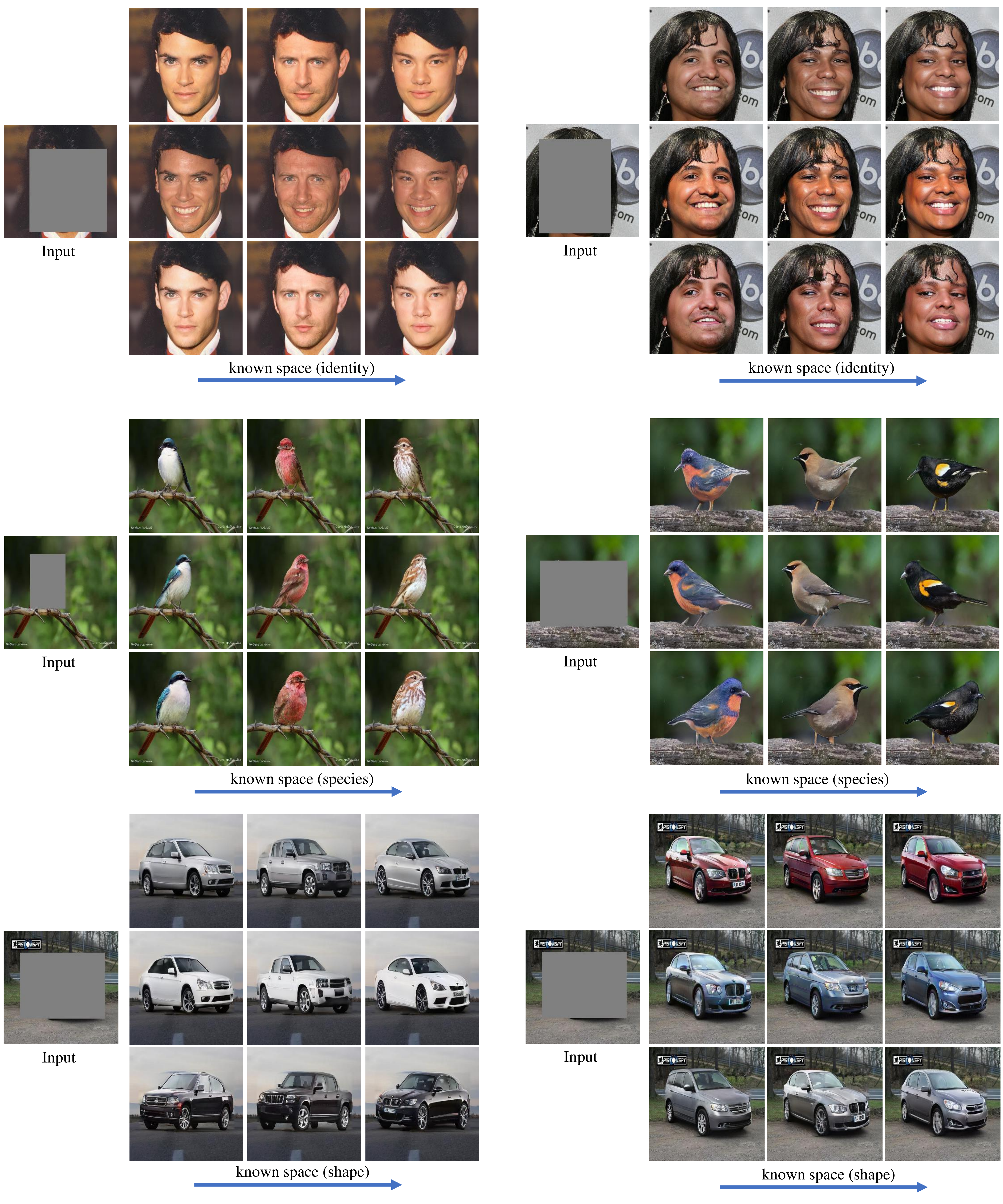}
    \caption{Disentanglement results on three different datasets.}
    \label{fig:supp_disentanglement}
\end{figure*}

\noindent\textbf{More qualitative results.} Finally, we show more results of comparisons with baselines in Figure~\ref{fig:baseline_face}, Figure~\ref{fig:baseline_bird} and Figure~\ref{fig:baseline_car}. Note that for CoModGAN~\cite{Zhao2021LargeSI} and CollageGAN~\cite{Li2021CollagingCG}, they have difficulties in changing object shape and pose for bird and car datasets. Figure~\ref{fig:supp_disentanglement} shows disentanglement results.

\end{document}